%% file: greenlightningai.tex
\documentclass[runningheads]{llncs}
\usepackage{geometry} 
\usepackage{amsfonts}
\usepackage{graphicx}
\usepackage{xspace}
\usepackage{hyperref}
\usepackage{subcaption}
\usepackage{color}
\usepackage{url}
\usepackage{booktabs}
\usepackage{tikz}
\usepackage{todonotes}
\usepackage{bbding}
\usepackage{multirow}
\usepackage{comment}
\usepackage{balance}
\usepackage{amsmath}
\usepackage[customcolors]{hf-tikz}
\usepackage{bm}
\usepackage[normalem]{ulem}
\usepackage{listings,multicol}
\usepackage{color}
\usepackage{xcolor}
\usepackage[edges]{forest}
\usetikzlibrary{calc,shadings,shapes.geometric}
\usepackage{pgfplots}

\lstdefinelanguage{Assembler}
{morekeywords={movq,movl,imull,vsubps,vaddps,vmulps,addi,vfsub,vfadd,vfmul,vs1r, vle32,vfmv,vsetvli,ldr,fsub,fmul,fadd,str,fmov},
sensitive=false,
morecomment=[l]{//},
morecomment=[s]{/*}{*/},
morestring=[b]",
}
\pgfplotsset{compat=1.15}

\newcommand{\glai}{\textrm{GreenLightningAI}\xspace}

\newcommand{\prettysmall}{\fontsize{6.7}{6.7}\selectfont}

\definecolor{gray98}{rgb}{0.93,0.93,0.93}
\definecolor{gray20}{rgb}{0.20,0.20,0.20}
\definecolor{gray25}{rgb}{0.25,0.25,0.25}
\definecolor{gray16}{rgb}{0.161,0.161,0.161}
\definecolor{gray60}{rgb}{0.6,0.6,0.6}
\definecolor{gray30}{rgb}{0.3,0.3,0.3}
\definecolor{bgray}{RGB}{248, 248, 248}
\definecolor{amgreen}{RGB}{40, 144, 40}
\definecolor{myblue}{RGB}{0, 40, 255}
\definecolor{amred}{RGB}{228,26,28}
\definecolor{amethyst}{rgb}{0.6, 0.4, 0.8}
\definecolor{mymauve}{rgb}{0.58,0,0.82}
\definecolor{LightGray}{gray}{0.9}

\def\addlegendimage{\csname pgfplots@addlegendimage\endcsname}
\pgfkeys{/pgfplots/number in legend/.style={
        /pgfplots/legend image code/.code={
            \node at (0.295,-0.0225){#1};
        },
        },
        /pgfplots/shape in legend/.style 2 args={
        /pgfplots/legend image code/.code={
            \node[draw,#1,fill=#2,minimum width=3mm] at (0.295,-0.0225){};
        },
        },
}
\input{figures/tikz_settings}

\begin{document}

\title{\glai:\\ An Efficient AI System with Decoupled\\ Structural and Quantitative Knowledge}
\titlerunning{An AI System with Decoupled Structural and Quantitative Knowledge}

\author{Jose Duato\inst{1} \and
Jose I. Mestre\inst{2} \and \\
Manuel~F.~Dolz\inst{2} \and
Enrique S. Quintana-Ortí\inst{3}}
\authorrunning{Jose Duato et al.}

\institute{Qsimov Quantum Computing S.L., Spain\\
\email{jduato@qsimov.com} \and
Universitat Jaume I de Castell\'o, Spain\\
\email{\{jmiravet,dolzm\}@uji.es} \and
Universitat Politècnica de Valencia, Spain \\
\email{quintana@disca.upv.es} 
}

\maketitle

\begin{abstract}
The number and complexity of artificial intelligence (AI) applications is growing relentlessly. As a result, even with the many algorithmic and mathematical advances experienced over past decades as well as the impressive energy efficiency and computational capacity of current hardware accelerators, training the most powerful and popular deep neural networks comes at very high economic and environmental costs.
Recognising that additional optimisations of conventional neural network training is very difficult, this work takes a radically different approach by proposing \glai, a new AI system design consisting of a linear model that is capable of emulating the behaviour of deep neural networks by subsetting the model for each particular sample. The new AI system stores the information required to select the system subset for a given sample (referred to as structural information) separately from the linear model parameters (referred to as quantitative knowledge). In this paper we present a proof of concept, showing that the structural information stabilises far earlier than the quantitative knowledge. Additionally, we show experimentally that the structural information can be kept unmodified when re-training the AI system with new samples while still achieving a validation accuracy similar to that obtained when re-training a neural network with similar size. Since the proposed AI system is based on a linear model, multiple copies of the model, trained with different datasets, can be easily combined. This enables faster and greener (re)-training algorithms, including incremental re-training and federated incremental re-training. 

\end{abstract}

\input{s1-intro}
\input{s2-related-work}

\input{s3-knowledge}
\input{s4-proofofconcept}

\input{s5-experiments}
\input{s6-greenlightningai}

\input{s7-remarks}

\section*{Acknowledgements}

This research was funded by the Qsimov Quantum Computing S.L. company through the Research and Development contract 10556/2022 ``Development of alternative techniques for the acceleration of deep neural network training'' with Universitat Jaume I. Manuel F. Dolz was also supported by the Plan Gen--T grant CIDEXG/2022/013 of the \emph{Generalitat Valenciana}. Jose I. Mestre was supported by the predoctoral grant ACIF/2021/281 of the \emph{Generalitat Valenciana}.

\bibliographystyle{IEEEtran}
\input{greenlightningai.bbl}

\end{document}

%% file: figures/tikz_settings.tex
\usepackage{tikz}
\usepackage{etoolbox} 
\usepackage{listofitems} 

\tikzset{>=latex} 
\usetikzlibrary{arrows, arrows.meta, calc, positioning, quotes, shapes}
\colorlet{mygray}{gray!30}
\colorlet{myred}{red!80!black}
\colorlet{myblue}{blue!80!black}
\colorlet{mygreen}{green!60!black}
\colorlet{mydarkred}{myred!40!black}
\colorlet{mydarkblue}{myblue!80!black}
\colorlet{mydarkgreen}{mygreen!40!black}
\tikzstyle{node}=[very thick,circle,draw=myblue,minimum size=22,inner sep=0.5,outer sep=0.6]
\tikzstyle{connect}=[->,very thick,mydarkblue,shorten >=0,sloped,midway,above]
\tikzstyle{connecd}=[->,very thick,myred,dashed,shorten >=0,sloped,midway,above]
\tikzset{ 
  node 1/.style={node,mydarkgreen,draw=mygreen,fill=mygreen!25},
  node 2/.style={node,mydarkblue,draw=myblue,fill=myblue!20},
  node 3/.style={node,mydarkred,draw=myred,fill=myred!20},
  relua/.pic = {
     \filldraw[fill=mygreen!25, draw=black] (-0.05,-0.05) rectangle (0.05,0.05);
     \path[draw=black,line width=1pt,-{Stealth[scale length=0.4,scale width=0.5]}] (-0.05,0) -- (0,0) -- (0.05,0.05);    
  },
  relui/.pic = {
     \filldraw[fill=myred!35, draw=black] (-0.05,-0.05) rectangle (0.05,0.05);
     \path[draw=black,line width=1pt,{Stealth[scale length=0.4,scale width=0.5]}-] (-0.05,0) -- (0,0) -- (0.05,0.05);     
  }  
}

\def\nstyle{int(\lay<\Nnodlen?min(2,\lay):3)} 
\def\nstyles{int(\lay<\Nnodlen?min(2,\lay):1)} 

\definecolor{fc}{HTML}{1E90FF}
\definecolor{h}{HTML}{228B22}
\definecolor{bias}{HTML}{87CEFA}
\definecolor{noise}{HTML}{8B008B}
\definecolor{conv}{HTML}{FFA500}
\definecolor{pool}{HTML}{B22222}
\definecolor{up}{HTML}{B22222}
\definecolor{view}{HTML}{FFFFFF}
\definecolor{bn}{HTML}{FFD700}
\definecolor{green_tab}{HTML}{d9ead3}
\definecolor{blue_tab}{HTML}{c9daf8}
\definecolor{yellow_tab}{HTML}{fff2cc}
\definecolor{color_samples}{HTML}{f4cccc}
\definecolor{color_pathselector}{HTML}{d9ead3}
\definecolor{color_actpattern}{HTML}{fff8c5}
\definecolor{color_estimator}{HTML}{d9d2e9}

\definecolor{color_base_pastelgreen}{rgb}{0.47, 0.87, 0.47}
\definecolor{color_base_pastelpink}{rgb}{1.0, 0.82, 0.86}
\definecolor{color_base_pastelviolet}{rgb}{0.8, 0.6, 0.79}
\definecolor{color_base_pastelyellow}{rgb}{0.99, 0.99, 0.59}
\definecolor{color_base_pastelred}{rgb}{1.0, 0.41, 0.38}
\definecolor{color_base_pastelgray}{rgb}{0.81, 0.81, 0.77}
\definecolor{color_base_green}{HTML}{009900}
\definecolor{color_base_blue}{HTML}{0000cc}
\definecolor{color_base_red}{HTML}{cc0000}
\colorlet{color_inputneuronfill}{color_base_green!30}
\colorlet{color_inputneuronborder}{color_base_green}
\colorlet{color_activeneuronfill}{color_base_blue!30}
\colorlet{color_activeneuronborder}{color_base_blue}
\colorlet{color_inactiveneuronfill}{color_base_red!30}
\colorlet{color_inactiveneuronborder}{color_base_red}
\colorlet{color_samples}{color_base_pastelred!50}
\colorlet{color_initialtraining}{color_base_pastelgray!30}
\colorlet{color_initialtrainingnetwork}{color_base_pastelgray!10}
\colorlet{color_pathselector}{color_base_pastelgreen!30}
\colorlet{color_pathselector_dark}{color_base_pastelgreen!70}
\colorlet{color_actpattern}{color_base_pastelyellow!50}
\colorlet{color_estimator}{color_base_pastelviolet!30}
\colorlet{color_estimator_dark}{color_base_pastelviolet!70}
\colorlet{color_estimator_light}{color_base_pastelviolet!10}

\tikzset{fc/.style={black,draw=fc,thick,fill=fc!30!white,rectangle,rounded corners=2.5,minimum height=0.6cm}}
\tikzset{sm/.style={black,draw=h,thick,fill=h!30!white,rectangle,rounded corners=2.5,minimum height=0.6cm}}
\tikzset{bias/.style={black,draw=black,thick,fill=bias,rectangle,rounded corners=2.5,minimum height=0.6cm}}
\tikzset{noise/.style={black,draw=noise,thick,fill=noise,rectangle,rounded corners=2.5,minimum height=0.6cm}}
\tikzset{conv/.style={black,draw=conv,thick,fill=conv!20!white,rectangle,rounded corners=2.5,minimum height=0.6cm}}
\tikzset{pool/.style={black,draw=red,thick,fill=red!30!white,rectangle,rounded corners=2.5,minimum height=0.6cm}}
\tikzset{up/.style={black,draw=black,thick,fill=up,rectangle,rounded corners=2.5,minimum height=0.6cm}}
\tikzset{input/.style={black,draw=black,thick,fill=view,rectangle,rounded corners=2.5,minimum height=0.6cm}}
\tikzset{bn/.style={black,draw=black,thick,fill=bn,rectangle,rounded corners,minimum height=0.6cm}}

\newcommand{\conv}{\textsf{Conv2D}\xspace}
\newcommand{\relu}{\textsf{ReLU}\xspace}

\newcommand{\bn}{\textsf{BatchNormalisation}\xspace}
\newcommand{\mpool}{\textsf{MaxPool2D}\xspace}
\newcommand{\apool}{\textsf{AveragePool2D}\xspace}
\newcommand{\fc}{\textsf{Fully-connected}\xspace}
\newcommand{\sm}{\textsf{Softmax}\xspace}

\pgfdeclarelayer{foreground}
\pgfdeclarelayer{background}
\pgfdeclarelayer{background1}
\pgfdeclarelayer{background2}
\pgfsetlayers{background2,background1,background,main,foreground}

%% file: s1-intro.tex
\section{Introduction}\label{sec:intro}
Deep learning (DL) has become a highly effective solution for addressing diverse challenges in image and speech recognition, natural language processing, and autonomous driving, among many others. Nevertheless, there still remain certain application domains where researchers, academia and companies are hesitant to leverage this powerful tool due to the significant time and energy costs necessary to train complex Deep Neural Networks (DNNs)~\cite{amodei2018ai}. The fact is that the severity of this problem is growing over time~\cite{nvidia2022h100}, even with the remarkable advances experienced thanks to specialised processor architectures and hardware accelerators in recent years~\cite{hennessy2019new,reuther2021ai}. 

The massive training costs associated with Deep Neural Networks (DNNs) stem from several factors. To achieve accurate recognition of meaningful and complex features, these DNNs must undergo extensive training using large datasets and sophisticated architectural models that feature an enormous number of neurons and, consequently, tunable parameters. Consequently, the training process for complex DNNs necessitates a significant number of arithmetic operations, using traditional Stochastic Gradient Descent (SGD)-based iterative method or any of its variants. This is due, in part, to challenges like vanishing gradients, which leads to slow convergence, and the implementation of techniques to avoid getting trapped in local minima. At this point, it is worth mentioning that the non-linearities embedded within DNNs play a crucial role in capturing real-world phenomena. However, in the case of large and complex DNNs, the successive adjustments required by iterative methods such as SGD to capture intricate non-linear relationships can lead to prolonged training times. As a result, this intensifies the computational demands and operational complexity involved in training such networks.

In addition to the previous discussion, for many applications, data evolves dynamically over time, necessitating periodic/non-periodic re-training. The consequence is that, since DNNs learn via some method that minimises a loss function, all the training samples must be processed every time the DNN is re-trained. (Otherwise, re-training usually leads to forgetting the contribution of the oldest samples.) Therefore, re-training is costly and difficult to implement without disruption, asking for a fundamentally different approach to that employed in DNN training. Equally significant to the lengthy computing time is the energy consumed during DNN re-training, particularly in the context of battery-operated devices such as cell phones and self-driving cars, that may benefit from re-training to adapt to the customer's behaviour and environmental changes, further underscoring the importance of addressing the energy consumption issue.

Focusing on the aforementioned problems, in this work we propose a new AI system design that consists of a linear model that is capable of emulating the piece-wise linear behaviour of DNNs that use the Rectified Linear Unit (ReLU) as the activation function. It does so by subsetting the linear model for each particular sample. The new AI system separately stores the information required to select the system subset for a given sample (referred to as structural information) from the linear model parameters (referred to as quantitative knowledge). This new system is designed with the goal of enabling simpler, faster and more environment-friendly (re)-training algorithms than the ones currently used for ReLU-based DNNs. This is achieved by the fact that multiple copies of a given linear model, trained with different datasets, can be easily combined. 

The key questions we address in this work are whether the structural information can be trained with lower cost than training a similarly-sized neural network, and whether the structural information can be kept unmodified when re-training the AI system without any significant loss in validation accuracy. This work focuses on answering those questions. In more detail, we make the following specific contributions:
\begin{itemize} 
\itemsep 0.1cm
\item  We revisit and define a number of known and new fundamental DNN concepts related to active paths and activation patterns that form the foundation of our new AI system.
\item We motivate our hypothesis that advocates for decoupling the structural and quantitative knowledge of a DNN. Basically, structural knowledge refers to the ability of the network to activate and deactivate paths, while quantitative knowledge refers to the numerical weight and bias values. A remarkable feature of the new AI system is that non-linearities are eliminated from the quantitative re-training part.
\item We validate our hypothesis through a proof-of-concept AI system that separates the structural and quantitative knowledge, comparing the evolution of both types of knowledge along the training process. We also evaluate the new AI system to re-train the LeNet-5, AlexNet and VGG8 convolutional neural networks (CNNs) on the CIFAR-10 dataset.
\item We present \glai, a new AI system that preserves the structural knowledge while updating only the quantitative knowledge with each re-training. One of the main properties of this system is that the quantitative part is transformed into a single layer without activation functions, thus containing only linear relationships and significantly reducing the cost of the re-training phase.
\end{itemize}

The remaining sections of the paper are organised as follows. Section~\ref{sec:related} provides a comprehensive overview of related work in the field. Section~\ref{sec:concepts} introduces several fundamental concepts and defines the notions of structural and quantitative knowledge. In Section~\ref{sec:proofofconcept}, we delve into the separation of structural and quantitative knowledge within DNNs. Section~\ref{sec:experimental} provides an evaluation and comparative analysis of the convergence of different DNNs using the knowledge-separated system, in contrast to the traditional approach. Section~\ref{sec:greenlightningai} unveils the internal structure of \glai.
Finally, Section~\ref{sec:remarks} concludes with some remarks and a brief discussion of avenues for future research.

%% file: s2-related-work.tex
\section{Related Work}\label{sec:related}
This section provides an overview of the state-of-the-art in alternative, non-iterative training algorithms and their application to machine learning. We also focus on related works that target challenges associated with DNN re-training, with a focus on efficient re-training methods and the incremental learning process.

Non-iterative algorithms, such as extreme machine learning~\cite{huang2006universal}, random vector functional link network~\cite{zhang2016comprehensive}, and neural networks with random weights~\cite{cao2018review}, have been proposed to analytically compute model parameters and reduce training costs. Those methods are in general computationally less expensive than the iterative solutions based on SGD which are mainstream in modern DL architectures~\cite{bianchi2017recurrent}. These approaches aim to alleviate the computational burden associated with iterative optimisation techniques commonly used in training DNNs. By analytically deriving the model parameters, these algorithms offer reduced training times. However, their applicability is restricted to non-DNN-based machine learning algorithms or shallow neural networks, which have limited capability in handling complex modelling tasks. 

Incremental learning is also an important task in many key applications with evolving data~\cite{klabjan2020neural}. Significant  examples arise in financial applications such as stock market forecasting~\cite{chatzis2018forecasting}, algorithmic trading~\cite{gerlein2016evaluating}, credit risk assessment~\cite{ma2019financial}, portfolio allocation~\cite{almahdi2017adaptive}, and asset pricing~\cite{sonksen2022machine}, as well as insurance risk assessment~\cite{kelley2018artificial}, preventive maintenance processes~\cite{dick2019deep}, and fine-tuning in natural language processing~\cite{tinn2023fine}. In these scenarios, the model needs to be periodically fine-tuned or re-trained using both old and new data. 

Transfer learning is a widely adopted technique employed when fine-tuning is necessary, enabling the establishment of a strong starting point for the DNN and significantly reducing training time. In this approach, a neural network is pre-trained for a different use case and dataset and, subsequently, its knowledge is leveraged and further refined using a specific dataset. This technique is particularly prevalent in image recognition, where the convolutional layers are first trained with a well-established dataset. Then, the fully-connected layers are fine-tuned for the desired task using a smaller dataset. Although transfer learning offers notable advantages, it generally yields lower accuracy when compared to training the entire network from scratch when a large dataset is available.

Another problem produced by transfer learning, as well as other re-training methods, is the phenomenon of catastrophic forgetting, which refers to the undesired loss of previously acquired knowledge~\cite{french1999catastrophic,kirkpatrick2017overcoming}. Several approaches have been proposed to tackle this problem, including memory replay, where re-training is performed by combining previous and new data or periodically re-training using all previous data~\cite{rolnick2019experience,van2020brain}; parameter adjustment based on importance~\cite{kirkpatrick2017overcoming}; and dynamic adjustment of the network architecture~\cite{yu2018slimmable,rusu2016progressive}. However, none of these approaches have been combined with the advancements in non-iterative training strategies found in the state-of-the-art.

In addition to the previous approaches, modular deep neural networks (MDNNs) propose an alternative where independent models, potentially pre-trained, process inputs and their outputs are combined through interface layers to produce a global response~\cite{auda1999modular,chen2015deep}. Ensembles of DNNs, which are a subset of MDNNs, typically process the same inputs to reach a final prediction~\cite{chen2015deep}. In these scenarios, re-training often involves adjusting the parameters of the last layer(s) while keeping the model components unchanged. Therefore, non-iterative re-training methods, such as linear regression approaches for one-layer feed-forward neural networks~\cite{castillo2002global}, can be feasible for these types of DNNs. Despite the superiority of MDNNs over DNNs, re-training individual models becomes necessary as global predictions become obsolete with new data.

In conclusion, there still remain important gaps in the design of efficient re-training methods for DNNs and in challenges of incremental learning such as catastrophic forgetting. This work proposes a radically different solution that separates the structural and quantitative knowledge of neural networks in order to circumvent the aforementioned problems.

%% file: s3-knowledge.tex
\section{Structural versus Quantitative Knowledge}\label{sec:concepts}
In this section, we first introduce a number of key concepts related to neural networks that lie at the foundations for our hypothesis that advocates for separating the structural knowledge from its quantitative counterpart.

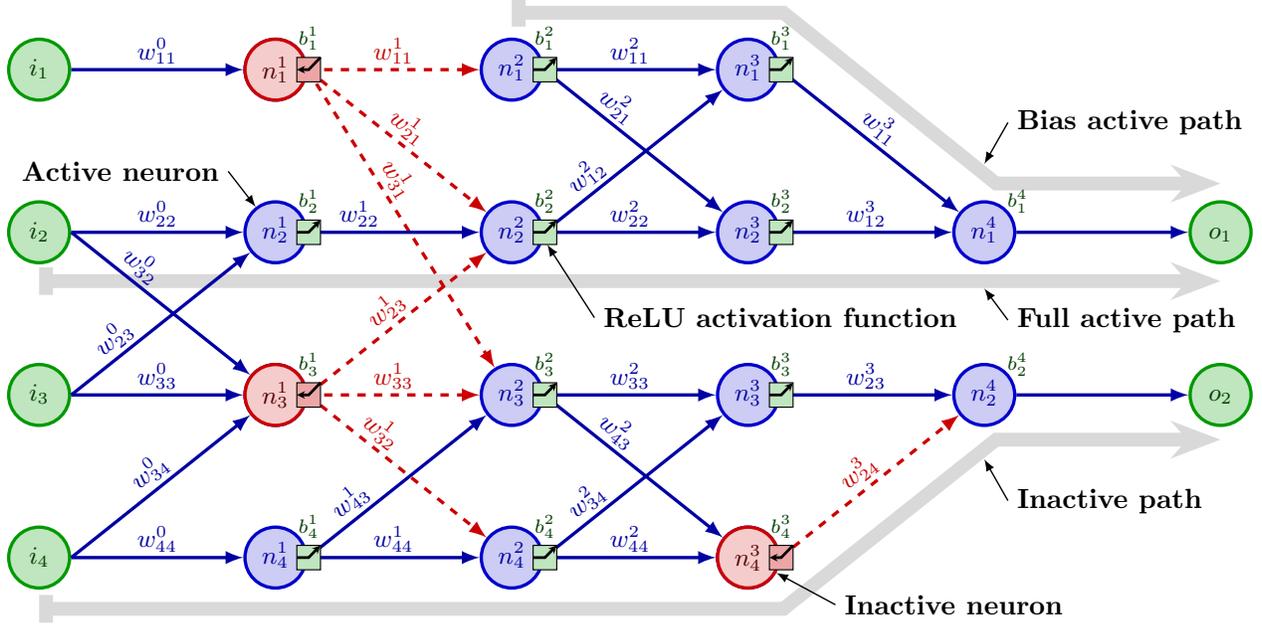
\begin{figure}[tbh!]
\centering
\resizebox{1.005\textwidth}{!}{\input{figures/nn}}
\vspace{-0.4cm}
\caption{Simple neural network with active/inactive neurons and paths. The neurons ($n$) are annotated with the layer number (superscript) and neuron identifier within the layer (subscript); the weights ($w$) are labelled with their input layer (superscript) and the output-input neuron identifiers (subscript); and the label for the bias ($b$) of each neuron follows the same scheme as the neuron. In this example, the blue neurons are active and all their outputs are active as well (marked with solid blue arrows). The red neurons and their outputs (marked with dashed red arrows) are inactive.}
\label{fig:nn}
\vspace{-0.3cm}
\end{figure}

\subsection{Fundamental concepts}
In the following, we utilise the simple neural network displayed in Figure~\ref{fig:nn} in order to define the following fundamental concepts:
\begin{itemize}
\itemsep 0.1cm
    \item A neuron computes a function, referred to as \textbf{activation function}, of the sum of the weighted neuron inputs plus a bias. The activation function is a non-linear mathematical function. The non-linearity enables the neural network to capture and learn complex relationships among the data. In the case of the ReLU activation function, $\sigma(x)=x, x>0$ and $\sigma(x)=0, x<=0$. Thus, the neuron output is equal to the weighted sum of its inputs (plus the bias) if that value is positive; otherwise, the output is zero. Note that the derivative of the ReLU activation function is $\sigma'(x)=1, x>0$ and $\sigma'(x)=0, x<0$. The new AI system presented in this work, as mentioned in the introduction, assumes ReLU as the activation function.

    \item An \textbf{active neuron}, for a given input sample, refers to a neuron for which the activation function produces a positive output, indicating its activation. An \textbf{inactive neuron} produces a zero output, indicating its inactivity. Note the relationship between the neuron activation status and the value of the derivative of the activation function. In the example, blue neurons are active while the rest, in red, are inactive.
    
    \item An \textbf{activation pattern} is defined as the set of neurons that are active for a given input sample~\cite{Hartmann21}. This pattern is determined by both the specific input sample and the values of the weights and biases within the neural network. In practice, samples with very similar features may produce the same activation pattern. In the simple neural network in the figure, the sample (values) in the input layer results in the activation pattern comprised by the neurons highlighted in blue.
    
    \item An \textbf{active path} is a sequence of active neurons in consecutive layers, with each active neuron being connected to the next  one in the path. The activation pattern for a specific input sample is defined by the collection of paths it activates. Given the interconnected nature of neural networks, the active paths may partially overlap. Additionally, each input-output neuron pair can be connected via multiple active paths.\\
    We distinguish between \textbf{full active paths}, which connect a specific input-output pair; and \textbf{bias active paths}, which connect a particular neuron (and not an input) to a given output. 
    In the example we can find the following four different types of paths (among many others):
    \emph{i)} two full paths that first overlap, later split, and finally join again (from $i_2$ to $o_1$); \emph{ii)} two full paths that first overlap, next split, later join, and finally overlap (from $i_4$ to $o_2$); \emph{iii)} an active bias path (from $n^2_1$ to $o_1$); and \emph{iv)} three groups of inactive paths (from $i_1$ to $o_1$, from $i_1$ to $o_2$, and from $i_3$ to $o_2$). 

    \item The \textbf{path weight} of a path is defined as the product of the weights along the path (including the bias of the source neuron in the product for bias paths). For a given sample, an output of a neural network can be expressed as the sum of contributions from all active paths leading to it, where each contribution is either the product of the corresponding path weight times the input value for a full path or just the path weight for a bias path. For example, the value of the output $o_1$ in the simple neural network in Figure~\ref{fig:nn} is given by:
\begin{equation*}
\resizebox{.82\textwidth}{!}{$\displaystyle
\begin{array}{rclclclcl}
 o_1 & = & b^4_1 \\ [0.1in]
     & + &  w^3_{11}  b^3_1 
       + \hfsetfillcolor{green!60!black!20}
       \hfsetbordercolor{green!60!black}
       \tikzmarkin{a}(0.05,0.-0.2)(-0.05,0.4)w^3_{11}  w^2_{11} b^2_1 \tikzmarkend{a} 
       + w^3_{11}  w^2_{12} b^2_2
       + w^3_{11}  w^2_{12} w^1_{22} b^1_2   
       + \hfsetfillcolor{blue!80!black!10}
       \hfsetbordercolor{blue!80!black}
       \tikzmarkin{b}(-0.32,-0.2)(-0.05,0.4)w^3_{11}  w^2_{12} w^1_{22} w^0_{22}\,i_2\tikzmarkend{b} 
       + w^3_{11}  w^2_{12} w^1_{22} w^0_{23} i_3\\[0.1in]
     & + & w^3_{12} b^3_2 
       + \hfsetfillcolor{green!60!black!20}
       \hfsetbordercolor{green!60!black}
       \tikzmarkin{c}(0.05,-0.22)(-0.05,0.4)w^3_{12} w^2_{21} b^2_1 \tikzmarkend{c}  
       + w^3_{12} w^2_{22} b^2_2   
       + w^3_{12} w^2_{22} w^1_{22} b^1_2   
       + \hfsetfillcolor{blue!80!black!10}
       \hfsetbordercolor{blue!80!black}
       \tikzmarkin{d}(-0.32,-0.22)(-0.05,0.4)w^3_{12} w^2_{22} w^1_{22} w^0_{22}\,i_2 \tikzmarkend{d} 
       + w^3_{12} w^2_{22} w^1_{22} w^0_{23} i_3. \\
\end{array}$}
\end{equation*}

In this expression, we can observe the contribution of the two full paths from $i_2$ to $o_1$, with the corresponding path weights obtained from the multiplication of the weights along the paths (marked with blue boxes) as well as those of the bias paths from $n_1^2$ to $o_1$ (marked with green boxes). It should also be noted that the above expression exposes that the inactive paths have no contribution to the output.
\end{itemize}

In general, the information in the neural network flows along the paths and, for a given sample, the associated output value corresponds to the sum of the contributions from the active paths leading to that output, regardless of whether or not those paths partially overlap. More precisely, the full paths model the contribution of the inputs to the outputs, while the bias paths model the contribution of the neuron biases to the outputs.

\subsection{Structural and quantitative knowledge}
\begin{figure}[ht!]
     \centering
     \begin{subfigure}[b]{0.48\textwidth}
         \centering
         \resizebox{1.0\textwidth}{!}{\input{figures/nn_structural}}
         \caption{Structural knowledge.}
         \label{fig:structural}
     \end{subfigure}
     \hfill
     \begin{subfigure}[b]{0.48\textwidth}
         \centering
         \resizebox{1.0\textwidth}{!}{\input{figures/nn_quantitative}}
         \caption{Quantitative knowledge.}
         \label{fig:quantitative}
     \end{subfigure}
\caption{Structural and quantitative knowledge for the example neural network in Figure~\ref{fig:nn}.}
\label{fig:structquant}
\end{figure}
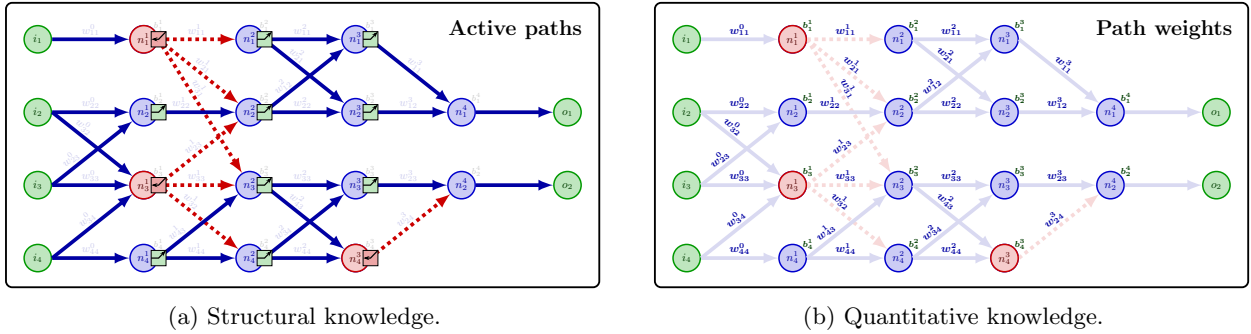

The new AI system introduced in this work is based on the hypothesis that the information or knowledge, learnt by a trained neural network, can be classified/separated into structural and quantitative. These two classes of knowledge are categorised as follows: 
\begin{itemize}
\itemsep 0.1cm
\item \textbf{Structural knowledge} refers to the information that the neural network has acquired in order to activate or deactivate a collection of paths, thereby generating specific activation patterns for specific input samples (see Figure~\ref{fig:structural}). It encompasses the network's learned ability to control the information flow through its layers via the activation functions, resulting in a distinctive activation pattern for each specific input sample. From a mathematical point of view, the structural knowledge can be represented by the values of the derivatives of the neuron activation functions for the different samples. The structural knowledge allows the network to effectively leverage its architecture to process and interpret samples in a manner that is aligned with the desired result.
\item \textbf{Quantitative knowledge}, in contrast, only encompasses the numerical values of the neuron weights and biases learnt by the neural network that allow it to generate accurate predictions for specific input samples. The quantitative knowledge does not involve the processing of activation functions (including their derivatives), since the active paths (and active neurons) are determined by the structural knowledge (see Figure~\ref{fig:quantitative}). The quantitative aspects collectively contribute to the numerical result produced by the neural network. In conclusion, when considering a set of fixed activations for a single sample, with some active and inactive neurons, the neural network can be regarded as a \textit{multivariable linear function}.
\end{itemize}

In general, combining structural and quantitative knowledge enables a neural network to effectively model complex non-linear systems. By decoupling the acquisition of structural knowledge from the adjustment of quantitative knowledge, it becomes possible to 1) design new systems that initially acquire the structural knowledge; 2) fine-tune the quantitative knowledge while avoiding the complexity of dealing with non-linearities; and 3) re-train multiple times by keeping the structural knowledge constant and updating only the quantitative knowledge.

%% file: figures/nn.tex
\begin{tikzpicture}[x=3cm,y=3.1cm]
  \readlist\Nnod{4,4,4,4,2,2} 
  \readlist\Nstr{n,m,k} 
  \readlist\Cstr{i,n^{\prev},o} 
  \def\yshift{0.75} 

  \draw [{Bar[width=10]}-{Stealth[scale length=0.7,scale width=0.7]},line width=5pt,mygray](1,-0.2) -- (6,-0.2);
  \path[draw=mygray,line width=5pt,{Bar[width=10]}-{Stealth[scale length=0.7,scale width=0.7]}] (3,0.9) -- (4.15,0.9) -- (5.05,0.2) -- (5.05,0.2) -- (6,0.2);  
  \path[draw=mygray,line width=5pt,{Bar[width=10]}-{Stealth[scale length=0.7,scale width=0.7]}] (1,2/3*-2-0.21) -- (4.15,2/3*-2-0.21) -- (5.05,-0.85) -- (6,-0.85);  

  \foreachitem \N \in \Nnod{
    \def\lay{\Ncnt} 
    \pgfmathsetmacro\prev{int(\Ncnt-1)} 
    \foreach \i [evaluate={\y=(\N/2-\i)*(2/3);
                           \x=\lay; \n=\nstyle;\ns=\nstyles;
                           \index=(\i);}] in {1,...,\N}{ 
      \node[node \ns] (N\lay-\i) at (\x,\y) {$\strut\Cstr[\n]_{\index}$};
      \ifnumcomp{\x}{>}{1}{
      \ifnumcomp{\x}{<}{6}{
      \node[anchor=north west,xshift=5,yshift=7,mydarkgreen] at (N\lay-\i.90) {\scriptsize $b^{\prev}_{\index}$};
      }{}}{}
    }
  }
  
  \draw[connect] (N1-1.360) -- (N2-1) node[midway,yshift=-1.5] {\small $w^0_{11}$};
  \draw[connect] (N1-2.360) -- (N2-2) node[midway,yshift=-1.5] {\small $w^0_{22}$};
  \draw[connect] (N1-2.360) -- (N2-3) node[midway,yshift=-1.5,pos=.33] {\small $w^0_{32}$};
  \draw[connect] (N1-3.360) -- (N2-2) node[midway,yshift=-1.5,pos=.3] {\small $w^0_{23}$};
  \draw[connect] (N1-3.360) -- (N2-3) node[midway,yshift=-1.5] {\small $w^0_{33}$};
  \draw[connect] (N1-4.360) -- (N2-3) node[midway,yshift=-1.5] {\small $w^0_{34}$};
  \draw[connect] (N1-4.360) -- (N2-4) node[midway,yshift=-1.5] {\small $w^0_{44}$};    
  
  \draw[connecd] (N2-1.360) -- (N3-1) node[midway,yshift=-1.5] {\small $w^1_{11}$};
  \draw[connecd] (N2-1.360) -- (N3-2) node[midway,yshift=-1.5] {\small $w^1_{21}$};
  \draw[connecd] (N2-1.360) -- (N3-3) node[midway,yshift=-1.5,pos=.4] {\small $w^1_{31}$};
  
  \draw[connect] (N2-2.360) -- (N3-2) node[midway,yshift=-1.5,pos=.3] {\small $w^1_{22}$};

  \draw[connecd] (N2-3.360) -- (N3-2) node[midway,yshift=-1.5] {\small $w^1_{23}$};
  \draw[connecd] (N2-3.360) -- (N3-3) node[midway,yshift=-1.5] {\small $w^1_{33}$};
  \draw[connecd] (N2-3.360) -- (N3-4) node[above,yshift=-1.5,pos=.35] {\small $w^1_{32}$};
  
  \draw[connect] (N2-4.360) -- (N3-3) node[above,yshift=-1.5,pos=.3] {\small $w^1_{43}$};
  \draw[connect] (N2-4.360) -- (N3-4) node[midway,yshift=-1.5] {\small $w^1_{44}$};
  
  \draw[connect] (N3-1.360) -- (N4-1) node[midway,yshift=-1.5] {\small $w^2_{11}$};
  \draw[connect] (N3-1.360) -- (N4-2) node[above,yshift=-1.5,pos=.35] {\small $w^2_{21}$};
  \draw[connect] (N3-2.360) -- (N4-1) node[above,yshift=-1.5,pos=.3] {\small $w^2_{12}$};
  \draw[connect] (N3-2.360) -- (N4-2) node[midway,yshift=-1.5] {\small $w^2_{22}$};
  \draw[connect] (N3-3.360) -- (N4-3) node[midway,yshift=-1.5] {\small $w^2_{33}$};
  \draw[connect] (N3-3.360) -- (N4-4) node[above,yshift=-1.5,pos=.35] {\small $w^2_{43}$};
  \draw[connect] (N3-4.360) -- (N4-3) node[above,yshift=-1.5,pos=.3] {\small $w^2_{34}$};
  \draw[connect] (N3-4.360) -- (N4-4) node[midway,yshift=-1.5] {\small $w^2_{44}$};  
  
  \draw[connect] (N4-1.360) -- (N5-1) node[midway,yshift=-1.5] {\small $w^3_{11}$};
  \draw[connect] (N4-2.360) -- (N5-1) node[midway,yshift=-1.5] {\small $w^3_{12}$};
  \draw[connect] (N4-3.360) -- (N5-2) node[midway,yshift=-1.5] {\small $w^3_{23}$};
  \draw[connecd] (N4-4.360) -- (N5-2) node[midway,yshift=-1.5] {\small $w^3_{24}$};    
  
  \draw[connect] (N5-1.360) -- (N6-1);
  \draw[connect] (N5-2.360) -- (N6-2);
  
  \node[node 3] (N2-1) at (2,2/3*1) {$n^1_1$};
  \node[node 3] (N2-3) at (2,2/3*-1) {$n^1_3$};
  \node[node 3] (N4-4) at (4,2/3*-2) {$n^3_4$};
  
  \draw(N2-1) [xshift=12] pic{relui};
  \draw(N2-2) [xshift=12] pic{relua};
  \draw(N2-3) [xshift=12] pic{relui};
  \draw(N2-4) [xshift=12] pic{relua}; 

  \draw(N3-1) [xshift=12] pic{relua};
  \draw(N3-2) [xshift=12] pic{relua};
  \draw(N3-3) [xshift=12] pic{relua};
  \draw(N3-4) [xshift=12] pic{relua}; 

  \draw(N4-1) [xshift=12] pic{relua};
  \draw(N4-2) [xshift=12] pic{relua};
  \draw(N4-3) [xshift=12] pic{relua};
  \draw(N4-4) [xshift=12] pic{relui};

  \draw [->,line width=0.5pt,black](1.8,0.25) -- (N2-2) node[anchor=east,pos=0] {\normalsize \textbf{Active neuron}};  
  \draw [->,line width=0.5pt,black](4.37, 2/3*-2.29) -- (N4-4) node[anchor=west,pos=0] {\normalsize \textbf{Inactive neuron}};
  \draw [->,line width=0.5pt,black](5.1, -1.1) -- (5,-0.93) node[anchor=west,pos=0] {\normalsize \textbf{Inactive path}};
  \draw [->,line width=0.5pt,black](5.1, -0.36) -- (5,-0.23) node[anchor=west,pos=0] {\normalsize \textbf{Full active path}};
  \draw [->,line width=0.5pt,black](5.1,  0.45) -- (5,0.28) node[anchor=west,pos=0] {\normalsize \textbf{Bias active path}};
  \draw [->,line width=0.5pt,black](3.35,-0.35) -- (3.05+0.10,-0.05) node[anchor=west,pos=0] {\normalsize \textbf{ReLU activation function}};

\end{tikzpicture}

%% file: figures/nn_structural.tex
\begin{tikzpicture}[x=3cm,y=3.1cm]

\tikzset{ 
  relua/.pic = {
     \filldraw[fill=mygreen!25, draw=black] (-0.07,-0.07) rectangle (0.07,0.07);
     \path[draw=black,line width=1pt,-{Stealth[scale length=0.7,scale width=0.6]}] (-0.07,0) -- (0,0) -- (0.07,0.07);    
  },
  relui/.pic = {
     \filldraw[fill=myred!35, draw=black] (-0.07,-0.07) rectangle (0.07,0.07);
     \path[draw=black,line width=1pt,{Stealth[scale length=0.7,scale width=0.6]}-] (-0.07,0) -- (0,0) -- (0.07,0.07);     
  }  
}
  \tikzstyle{connect}=[->,line width=3pt,mydarkblue,shorten >=0,sloped,midway,above]
  \tikzstyle{connecd}=[->,line width=3pt,myred,dashed,shorten >=0,sloped,midway,above]
  \readlist\Nnod{4,4,4,4,2,2} 
  \readlist\Nstr{n,m,k} 
  \readlist\Cstr{i,n^{\prev},o} 
  \def\yshift{0.75} 

  
  \foreachitem \N \in \Nnod{
    \def\lay{\Ncnt} 
    \pgfmathsetmacro\prev{int(\Ncnt-1)} 
    \foreach \i [evaluate={\y=(\N/2-\i)*(2/3);
                           \x=\lay; \n=\nstyle;\ns=\nstyles;
                           \index=(\i);}] in {1,...,\N}{ 
      \node[node \ns] (N\lay-\i) at (\x,\y) {$\strut\Cstr[\n]_{\index}$};
      \ifnumcomp{\x}{>}{1}{
      \ifnumcomp{\x}{<}{6}{
      \node[anchor=north west,xshift=5,yshift=7,mydarkgreen] at (N\lay-\i.90) {\scriptsize \color{black!20}$b^{\prev}_{\index}$};
      }{}}{}
    }
  }
  
  \draw[connect] (N1-1.360) -- (N2-1) node[midway,yshift=-1.5] {\small \color{mydarkblue!20}$w^0_{11}$};
  \draw[connect] (N1-2.360) -- (N2-2) node[midway,yshift=-1.5] {\small \color{mydarkblue!20}$w^0_{22}$};
  \draw[connect] (N1-2.360) -- (N2-3) node[midway,yshift=-1.5,pos=.33] {\small \color{mydarkblue!20}$w^0_{32}$};
  \draw[connect] (N1-3.360) -- (N2-2) node[midway,yshift=-1.5,pos=.3] {\small \color{mydarkblue!20}$w^0_{23}$};
  \draw[connect] (N1-3.360) -- (N2-3) node[midway,yshift=-1.5] {\small \color{mydarkblue!20}$w^0_{33}$};
  \draw[connect] (N1-4.360) -- (N2-3) node[midway,yshift=-1.5] {\small \color{mydarkblue!20}$w^0_{34}$};
  \draw[connect] (N1-4.360) -- (N2-4) node[midway,yshift=-1.5] {\small \color{mydarkblue!20}$w^0_{44}$};

  \draw[connecd] (N2-1.360) -- (N3-1) node[midway,yshift=-1.5] {\small \color{mydarkblue!20}$w^1_{11}$};
  \draw[connecd] (N2-1.360) -- (N3-2) node[midway,yshift=-1.5] {\small \color{mydarkblue!20}$w^1_{21}$};
  \draw[connecd] (N2-1.360) -- (N3-3) node[midway,yshift=-1.5,pos=.4] {\small \color{mydarkblue!20}$w^1_{31}$};
  
  \draw[connect] (N2-2.360) -- (N3-2) node[midway,yshift=-1.5,pos=.3] {\small \color{mydarkblue!20}$w^1_{22}$};

  \draw[connecd] (N2-3.360) -- (N3-2) node[midway,yshift=-1.5] {\small \color{mydarkblue!20}$w^1_{23}$};
  \draw[connecd] (N2-3.360) -- (N3-3) node[midway,yshift=-1.5] {\small \color{mydarkblue!20}$w^1_{33}$};
  \draw[connecd] (N2-3.360) -- (N3-4) node[above,yshift=-1.5,pos=.35] {\small \color{mydarkblue!20}$w^1_{32}$};
  
  \draw[connect] (N2-4.360) -- (N3-3) node[above,yshift=-1.5,pos=.3] {\small \color{mydarkblue!20}$w^1_{43}$};
  \draw[connect] (N2-4.360) -- (N3-4) node[midway,yshift=-1.5] {\small \color{mydarkblue!20}$w^1_{44}$};
  
  \draw[connect] (N3-1.360) -- (N4-1) node[midway,yshift=-1.5] {\small \color{mydarkblue!20}$w^2_{11}$};
  \draw[connect] (N3-1.360) -- (N4-2) node[above,yshift=-1.5,pos=.35] {\small \color{mydarkblue!20}$w^2_{21}$};
  \draw[connect] (N3-2.360) -- (N4-1) node[above,yshift=-1.5,pos=.3] {\small \color{mydarkblue!20}$w^2_{12}$};
  \draw[connect] (N3-2.360) -- (N4-2) node[midway,yshift=-1.5] {\small \color{mydarkblue!20}$w^2_{22}$};
  \draw[connect] (N3-3.360) -- (N4-3) node[midway,yshift=-1.5] {\small \color{mydarkblue!20}$w^2_{33}$};
  \draw[connect] (N3-3.360) -- (N4-4) node[above,yshift=-1.5,pos=.35] {\small \color{mydarkblue!20}$w^2_{43}$};
  \draw[connect] (N3-4.360) -- (N4-3) node[above,yshift=-1.5,pos=.3] {\small \color{mydarkblue!20}$w^2_{34}$};
  \draw[connect] (N3-4.360) -- (N4-4) node[midway,yshift=-1.5] {\small \color{mydarkblue!20}$w^2_{44}$};  
  
  \draw[connect] (N4-1.360) -- (N5-1) node[midway,yshift=-1.5] {\small \color{mydarkblue!20}$w^3_{11}$};
  \draw[connect] (N4-2.360) -- (N5-1) node[midway,yshift=-1.5] {\small \color{mydarkblue!20}$w^3_{12}$};
  \draw[connect] (N4-3.360) -- (N5-2) node[midway,yshift=-1.5] {\small \color{mydarkblue!20}$w^3_{23}$};
  \draw[connecd] (N4-4.360) -- (N5-2) node[midway,yshift=-1.5] {\small \color{mydarkblue!20}$w^3_{24}$};    
  
  \draw[connect] (N5-1.360) -- (N6-1);
  \draw[connect] (N5-2.360) -- (N6-2);
  
  \node[node 3] (N2-1) at (2,2/3*1) {$n^1_1$};
  \node[node 3] (N2-3) at (2,2/3*-1) {$n^1_3$};
  \node[node 3] (N4-4) at (4,2/3*-2) {$n^3_4$};
  
  \draw(N2-1) [xshift=12] pic{relui};
  \draw(N2-2) [xshift=12] pic{relua};
  \draw(N2-3) [xshift=12] pic{relui};
  \draw(N2-4) [xshift=12] pic{relua}; 

  \draw(N3-1) [xshift=12] pic{relua};
  \draw(N3-2) [xshift=12] pic{relua};
  \draw(N3-3) [xshift=12] pic{relua};
  \draw(N3-4) [xshift=12] pic{relua}; 

  \draw(N4-1) [xshift=12] pic{relua};
  \draw(N4-2) [xshift=12] pic{relua};
  \draw(N4-3) [xshift=12] pic{relua};
  \draw(N4-4) [xshift=12] pic{relui};
  
  \draw [line width=1.5pt,rounded corners=5] (0.7,-1.6) rectangle (6.3,1.0);
  \node [above=1.3,xshift=-42] at (N6-2) {\LARGE \textbf{Active paths}}; 
 
\end{tikzpicture}

%% file: figures/nn_quantitative.tex
\begin{tikzpicture}[x=3cm,y=3.1cm]
  \tikzstyle{connect}=[->,line width=3pt,mydarkblue!15,shorten >=0,sloped,midway,above]
  \tikzstyle{connecd}=[->,line width=3pt,myred!15,dashed,shorten >=0,sloped,midway,above]
  \readlist\Nnod{4,4,4,4,2,2} 
  \readlist\Nstr{n,m,k} 
  \readlist\Cstr{i,n^{\prev},o} 
  \def\yshift{0.75} 

  
  \foreachitem \N \in \Nnod{
    \def\lay{\Ncnt} 
    \pgfmathsetmacro\prev{int(\Ncnt-1)} 
    \foreach \i [evaluate={\y=(\N/2-\i)*(2/3);
                           \x=\lay; \n=\nstyle;\ns=\nstyles;
                           \index=(\i);}] in {1,...,\N}{ 
      \node[node \ns] (N\lay-\i) at (\x,\y) {$\strut\Cstr[\n]_{\index}$};
      \ifnumcomp{\x}{>}{1}{
      \ifnumcomp{\x}{<}{6}{
      \node[anchor=north west,xshift=5,yshift=7,mydarkgreen] at (N\lay-\i.90) {\scriptsize $\bm{b^{\prev}_{\index}}$};
      }{}}{}
    }
  }
  
  \draw[connect] (N1-1.360) -- (N2-1) node[midway,yshift=-1.5] {\small \color{mydarkblue}$\bm{w^0_{11}}$};
  \draw[connect] (N1-2.360) -- (N2-2) node[midway,yshift=-1.5] {\small \color{mydarkblue}$\bm{w^0_{22}}$};
  \draw[connect] (N1-2.360) -- (N2-3) node[midway,yshift=-1.5,pos=.33] {\small \color{mydarkblue}$\bm{w^0_{32}}$};
  \draw[connect] (N1-3.360) -- (N2-2) node[midway,yshift=-1.5,pos=.3] {\small \color{mydarkblue}$\bm{w^0_{23}}$};
  \draw[connect] (N1-3.360) -- (N2-3) node[midway,yshift=-1.5] {\small \color{mydarkblue}$\bm{w^0_{33}}$};
  \draw[connect] (N1-4.360) -- (N2-3) node[midway,yshift=-1.5] {\small \color{mydarkblue}$\bm{w^0_{34}}$};
  \draw[connect] (N1-4.360) -- (N2-4) node[midway,yshift=-1.5] {\small \color{mydarkblue}$\bm{w^0_{44}}$};

  \draw[connecd] (N2-1.360) -- (N3-1) node[midway,yshift=-1.5] {\small \color{mydarkblue}$\bm{w^1_{11}}$};
  \draw[connecd] (N2-1.360) -- (N3-2) node[midway,yshift=-1.5] {\small \color{mydarkblue}$\bm{w^1_{21}}$};
  \draw[connecd] (N2-1.360) -- (N3-3) node[midway,yshift=-1.5,pos=.4] {\small \color{mydarkblue}$\bm{w^1_{31}}$};
  
  \draw[connect] (N2-2.360) -- (N3-2) node[midway,yshift=-1.5,pos=.3] {\small \color{mydarkblue}$\bm{w^1_{22}}$};

  \draw[connecd] (N2-3.360) -- (N3-2) node[midway,yshift=-1.5] {\small \color{mydarkblue}$\bm{w^1_{23}}$};
  \draw[connecd] (N2-3.360) -- (N3-3) node[midway,yshift=-1.5] {\small \color{mydarkblue}$\bm{w^1_{33}}$};
  \draw[connecd] (N2-3.360) -- (N3-4) node[above,yshift=-1.5,pos=.35] {\small \color{mydarkblue}$\bm{w^1_{32}}$};
  
  \draw[connect] (N2-4.360) -- (N3-3) node[above,yshift=-1.5,pos=.3] {\small \color{mydarkblue}$\bm{w^1_{43}}$};
  \draw[connect] (N2-4.360) -- (N3-4) node[midway,yshift=-1.5] {\small \color{mydarkblue}$\bm{w^1_{44}}$};
  
  \draw[connect] (N3-1.360) -- (N4-1) node[midway,yshift=-1.5] {\small \color{mydarkblue}$\bm{w^2_{11}}$};
  \draw[connect] (N3-1.360) -- (N4-2) node[above,yshift=-1.5,pos=.35] {\small \color{mydarkblue}$\bm{w^2_{21}}$};
  \draw[connect] (N3-2.360) -- (N4-1) node[above,yshift=-1.5,pos=.3] {\small \color{mydarkblue}$\bm{w^2_{12}}$};
  \draw[connect] (N3-2.360) -- (N4-2) node[midway,yshift=-1.5] {\small \color{mydarkblue}$\bm{w^2_{22}}$};
  \draw[connect] (N3-3.360) -- (N4-3) node[midway,yshift=-1.5] {\small \color{mydarkblue}$\bm{w^2_{33}}$};
  \draw[connect] (N3-3.360) -- (N4-4) node[above,yshift=-1.5,pos=.35] {\small \color{mydarkblue}$\bm{w^2_{43}}$};
  \draw[connect] (N3-4.360) -- (N4-3) node[above,yshift=-1.5,pos=.3] {\small \color{mydarkblue}$\bm{w^2_{34}}$};
  \draw[connect] (N3-4.360) -- (N4-4) node[midway,yshift=-1.5] {\small \color{mydarkblue}$\bm{w^2_{44}}$};  
  
  \draw[connect] (N4-1.360) -- (N5-1) node[midway,yshift=-1.5] {\small \color{mydarkblue}$\bm{w^3_{11}}$};
  \draw[connect] (N4-2.360) -- (N5-1) node[midway,yshift=-1.5] {\small \color{mydarkblue}$\bm{w^3_{12}}$};
  \draw[connect] (N4-3.360) -- (N5-2) node[midway,yshift=-1.5] {\small \color{mydarkblue}$\bm{w^3_{23}}$};
  \draw[connecd] (N4-4.360) -- (N5-2) node[midway,yshift=-1.5] {\small \color{mydarkblue}$\bm{w^3_{24}}$};    
  
  \draw[connect] (N5-1.360) -- (N6-1);
  \draw[connect] (N5-2.360) -- (N6-2);
  
  \node[node 3] (N2-1) at (2,2/3*1) {$n^1_1$};
  \node[node 3] (N2-3) at (2,2/3*-1) {$n^1_3$};
  \node[node 3] (N4-4) at (4,2/3*-2) {$n^3_4$};
  
  \draw [line width=1.5pt,rounded corners=5] (0.7,-1.6) rectangle (6.3,1.0);
  \node [above=1.3,xshift=-42] at (N6-2) {\LARGE \textbf{Path weights}}; 
 
\end{tikzpicture}

%% file: s4-proofofconcept.tex
\section{Proof-of-Concept}\label{sec:proofofconcept}

This section presents a proof-of-concept AI system that decouples the structural and quantitative knowledge within a DNN, enabling us to re-train one kind of knowledge while keeping the other one unmodified. We outline the key stages of the re-training process that updates only the quantitative knowledge. It is important to note that the system discussed in this section does not represent our design of the GreenLightingAI system. Instead, we tried to keep a structure as close as possible to a DNN, also implementing a re-training mechanism closely resembling the one for the original DNN. This approach aims at validating the hypothesis that the two types of knowledge can indeed be decoupled.

\subsection{System Definition}
The goal is to reduce the amount of computations required to re-train an AI system. We intend to achieve this goal by decoupling structural and quantitative knowledge, and preserving the structural knowledge while re-training only the quantitative information. This is fundamentally different from traditional training, for which both types of knowledge are updated during the re-training process. 

The new approach is viable because we found a way to decouple structural from quantitative knowledge and, as proved later, the structural knowledge exhibits a high degree of stability across successive re-training operations with additional samples~\cite{Hartmann21}. The reason for this behaviour lies in that the high-level features of the samples are typically captured in the early stages of the training process, while the refinement of the low-level features occurs in later stages, closer to the end of the learning process. The stabilisation of the activation patterns, which represent the structural knowledge of the neural network, is thus an indication that the high-level features have been effectively learned. In contrast, the fine-tuning of weights and biases, constituting the quantitative knowledge, enables the network to minimise errors, for example in the loss function, once the activation patterns (active paths) are mostly stabilised. 

From a different point of view, it is worth taking into account that the structural knowledge (i.e., the set of active paths for each sample) can be represented by the values of the derivatives of the neuron activation functions for the different samples. For a given sample, a derivative equal to zero means that the corresponding neuron is inactive, while an active neuron has a derivative equal to one. Moreover, the derivative of the ReLU is constant regardless of any variations in the value of the sum of the weighted neuron inputs (plus bias) as long as the sign of the sum does not change. Thus, small weight variations are unlikely to change the sign of the sum. Since neuron weight variations at successive iterations become progressively smaller as the training process converges, it implies that those small variations are less likely to affect the structural knowledge, thus confirming its high degree of stability.

Starting from a given neural network, the proof-of-concept AI system builds two copies of such network. The first copy is used for computing the derivatives of the neuron activation functions and it is not updated when re-training the system. Those derivatives indicate the activation status of the corresponding neurons, and sequences of active neurons form active paths. Hence, this copy will be referred to as the \textbf{path selector}. The second copy uses the derivatives computed by the first copy both for inference and re-training (during the backward pass) and its tunable parameters are updated when re-training. Since the second copy is used to estimate the system output values, it will be referred to as the \textbf{estimator}. We will later reuse both concepts (i.e. path selector and estimator) when designing the GreenLightingAI system.

Thus, the proof-of-concept AI system consists of the following two modules:
\begin{itemize}
\itemsep 0.1cm
\item The \textbf{path selector} contains the structural knowledge and is responsible for obtaining the set of active paths for a given sample. Afterwards, it feeds this information to the estimator. 
 
\item The \textbf{estimator} contains the quantitative knowledge and is responsible for delivering accurate inference results (i.e., a prediction) for the input sample by selecting and using a subset of parameters according to the neuron activation pattern dictated by the path selector.
\end{itemize}

\subsection{Implementation}
In this subsection, we offer a comprehensive description of the experiments conducted during the initial setup and subsequent re-training iterations with additional samples. Figure~\ref{fig:retraining} illustrates the workflow devised for this proof-of-concept AI system, delineated into the following four phases:

\begin{figure}[ht!]
\centering
\includegraphics[width=0.95\columnwidth]{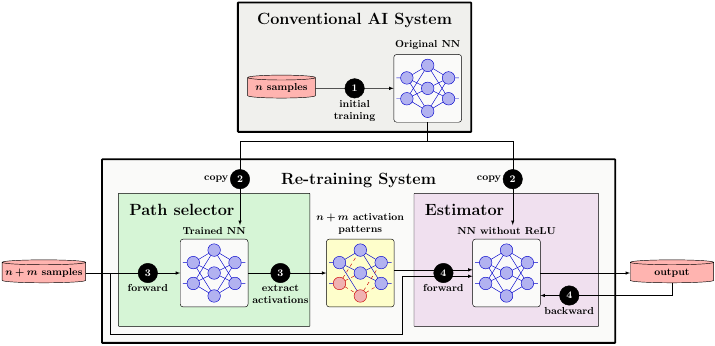}
\caption{Diagram of the proposed AI system, including the initial training, path selector and the estimator.}
\label{fig:retraining}
\end{figure}

\begin{itemize}
\itemsep 0.1cm    
\item \textbf{Phase 1: Initial training of the original neural network.} \label{phase1}
During this phase, the network is initialised; in the case of re-training, the previous version of the network serves as the starting point. For a newly initialised network, the initial training involves using a data subset comprising, for example, $n$ samples, typically with the assistance of an optimiser such as SGD. It is crucial to carefully select the data subset and the number of training epochs to stabilise the structural information in the path selector, ensuring reasonably accurate predictions at the end of the re-training process. Throughout this initial training process, the neural network develops the ability to recognise high-level features and distinguish among different samples. 

\item \textbf{Phase 2: Initialisation of the path selector and the estimator.}
After completing the initial training, the path selector receives a copy of the trained neural network. The parameters of this copy remain constant during subsequent re-trainings unless the structural knowledge becomes obsolete, in which case the network needs to be trained again as per Phase 1. Also, after the initial training, the estimator receives a version of the trained neural network without the activation functions (nonlinearities). 

\item \textbf{Phase 3: Preparation of re-training: Obtaining the activation patterns from the path selector.}
Prior to the re-training process of the estimator (Phase 4), it is necessary to obtain the activation patterns associated with all the re-training samples, including the original $n$ inputs plus the (say) $m$ new ones. These activation patterns are obtained by performing inference (forward pass) using the static copy of the neural network included in the path selector. During this inference process, the nonlinear functions of the neural network will activate differently for each sample, thereby generating distinct activation patterns.

\item \textbf{Phase 4: Re-training the estimator with additional samples.}
In this phase, the estimator undergoes a quantitative knowledge-only re-training of the neural network using the expanded set of $n+m$ samples using the activation patterns for those samples computed by the path selector in Phase~3. 

During the re-training process, the following operations are performed for each batch of samples. First, the estimator performs a forward pass. At each layer, the activation pattern provided by the path selector for each sample determines whether a neuron output should be active or not (regardless of the value of the sum of the weighted inputs plus the bias). Next, the loss function is used to compute the deviation with respect to the ground truth. In the backward pass, the gradients are propagated backwards as usual, except that the derivatives of the activation functions are taken from the corresponding activation pattern provided by the path selector. Remember that, for each sample, the activation status of each neuron is equivalent to the derivative of the activation function. Finally, the estimator weights are re-adjusted using the SGD optimiser.
\end{itemize}

This re-training process implements the fundamental concepts of the new AI system, while being similar to the conventional DNN re-training process. In practice, this re-training can be significantly faster since the estimator does not incorporate nonlinear activation functions.

%% file: s5-experiments.tex
\section{Hypothesis Validation}\label{sec:experimental}

In this section, we evaluate the previously described proof-of-concept training compared to traditional training through a series of experiments, analysing the accuracy of the different DNN models. 

\subsection{Proof-of-concept implementation and libraries}
The proof-of-concept implementation of the proof-of-concept AI system involves reformulating the layers typically used in CNNs. The aim is to track the activation patterns for each sample as it passes through the network. This is achieved by using binary masks that store the activation patterns and pass them on to the estimator during the re-training phase. In this phase, only the weights and biases of the active neurons are updated. 

The implementation leverages basic codes that incorporate the forward and backward methods from the PyDTNN training framework~\cite{PyDTNN2}. TensorFlow v2.6.2 is used for the initialisation. The implementation also relies on the Numpy and Pandas packages in Python to harness the performance of multi-CPU functionality.

\subsection{Methodology}
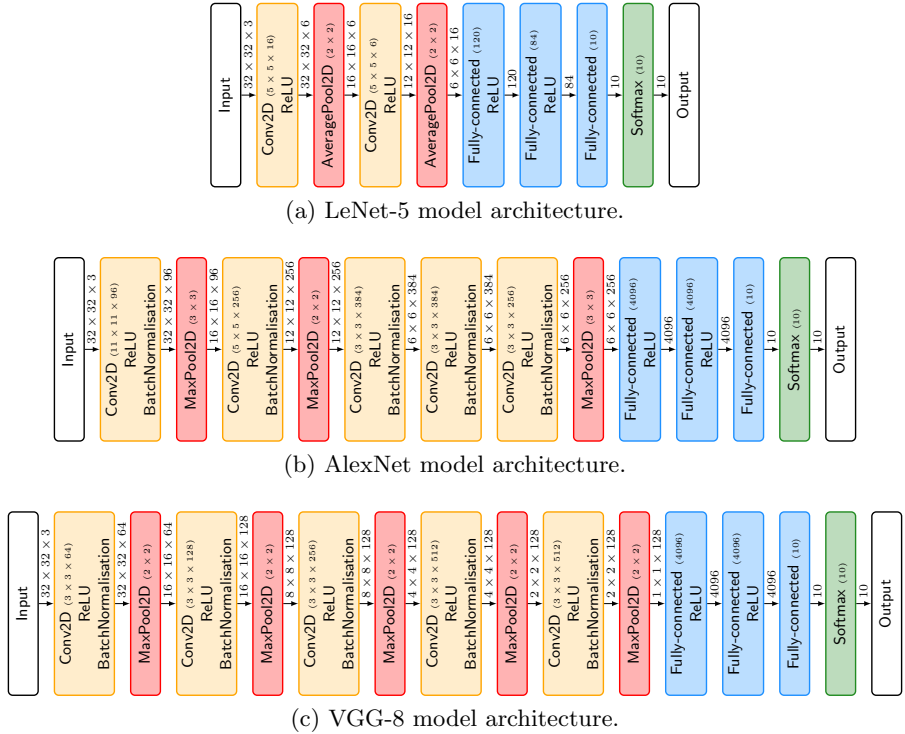
\begin{figure}[ht!]
\centering
     \begin{subfigure}[b]{1.0\textwidth}
         \centering
         \resizebox{!}{2.6cm}{\input{figures/lenet5}}\vspace{-0.15cm}
         \caption{LeNet-5 model architecture.}
         \label{fig:lenet5}
     \end{subfigure}
     \\[0.3cm]
     \begin{subfigure}[b]{1.0\textwidth}
         \centering
         \resizebox{!}{2.6cm}{\input{figures/alexnet}}\vspace{-0.15cm}
         \caption{AlexNet model architecture.}
         \label{fig:alexnet}
     \end{subfigure}
     \\[0.3cm]
     \begin{subfigure}[b]{1.0\textwidth}
         \centering
         \resizebox{!}{2.6cm}{\input{figures/vgg8}}\vspace{-0.15cm}
         \caption{VGG-8 model architecture.}
         \label{fig:vgg8}
     \end{subfigure}     
\caption{DNN model architectures.}\vspace{-0.4cm}
\label{fig:models}
\end{figure}

To validate the proposed re-training system, we conducted several experiments using three well-known neural networks: (a) LeNet-5, (b) AlexNet, and (c) VGG8, on the CIFAR-10 dataset (see Figure~\ref{fig:models}). Initially, each of these models was pre-trained using the traditional training approach for 200 epochs, using $n=$1,024, 1,024 and 512 samples, respectively. \textcolor{black}{Additionally, in order to analyse the impact of the quality of the path selector on the accuracy of the estimator after the re-training process, the same experiments were repeated for the same models, pre-trained with $4,096$ samples.}
These samples were randomly selected from the aforementioned dataset. 

For each subsequent re-training experiment with an increasing number of additional samples ($m$), our proof-of-concept AI system was initialised by creating two copies of the network: one for the path selector and another one for the estimator. The path selector performed inference using the set of $n+m$ samples and recorded the activation pattern for each sample. Subsequently, the estimator re-trained the quantitative knowledge of the neural network, specifically the weights and biases associated with the active paths for each of the $n+m$ samples, using the SGD optimiser for 50 epochs. It is important to note that all trainable layers, including convolutional and fully-connected layers, as well as batch normalisation layers, were trained. Furthermore, the neural network copy used by the estimator did not include activation functions. Instead, it utilised the activation patterns obtained from the path selector to determine the active and inactive neurons for each sample.

To compare the results of our proof-of-concept AI system with traditional DNNs in consecutive re-training experiments with an increasing number of samples, we also re-trained the same models using the traditional approach, training them with $n+m$ samples for 50 epochs. This allows us to assess the effectiveness of our approach compared to the conventional training method.

\subsection{Structural knowledge stabilisation analysis}
One of our hypotheses is that structural knowledge stabilises during the early stages of the training process. This enables the preservation of structural knowledge while re-training only the quantitative information. To validate this hypothesis, we compare the variations in activation patterns throughout the training process.

\begin{figure}[ht!]
\centering
\includegraphics[width=1\columnwidth]{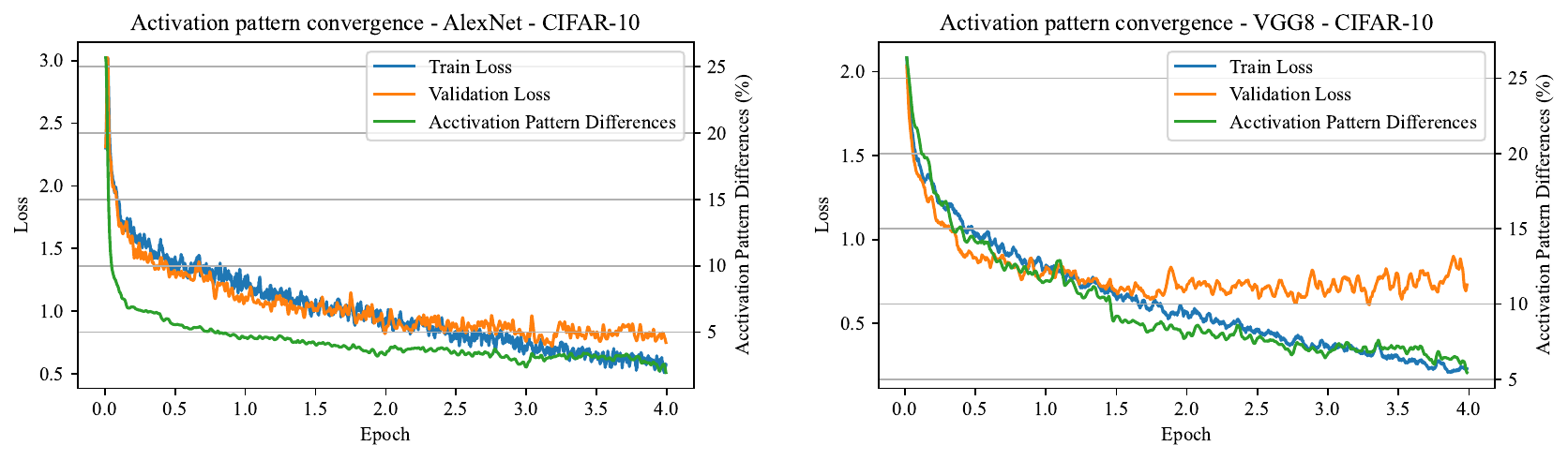}\\[0.1cm]
\caption{Activation pattern convergence for AlexNet and VGG-8 using CIFAR-10.}
\label{fig:activationpatternconvergence}
\end{figure}

Figure~\ref{fig:activationpatternconvergence} illustrates these variations alongside the training and validation loss at each training step for both the AlexNet and VGG-8 networks. The variations are assessed via the difference between active and inactive paths for each sample in the validation dataset, quantifying the percentage of paths that have changed compared to the state of the previously trained neural network, using the same random seed for weight initialisation and batch shuffling. As evident in the plot, the activation pattern variations stabilise much earlier than the validation loss, even when the weights are still being updated. Therefore, we can confidently assert that structural knowledge stabilises within a few epochs, while the quantitative knowledge remains amenable to updates.

\subsection{Quantitative knowledge analysis with successive re-trainings}
\begin{figure}[ht!]
\centering
\includegraphics[width=1\columnwidth]{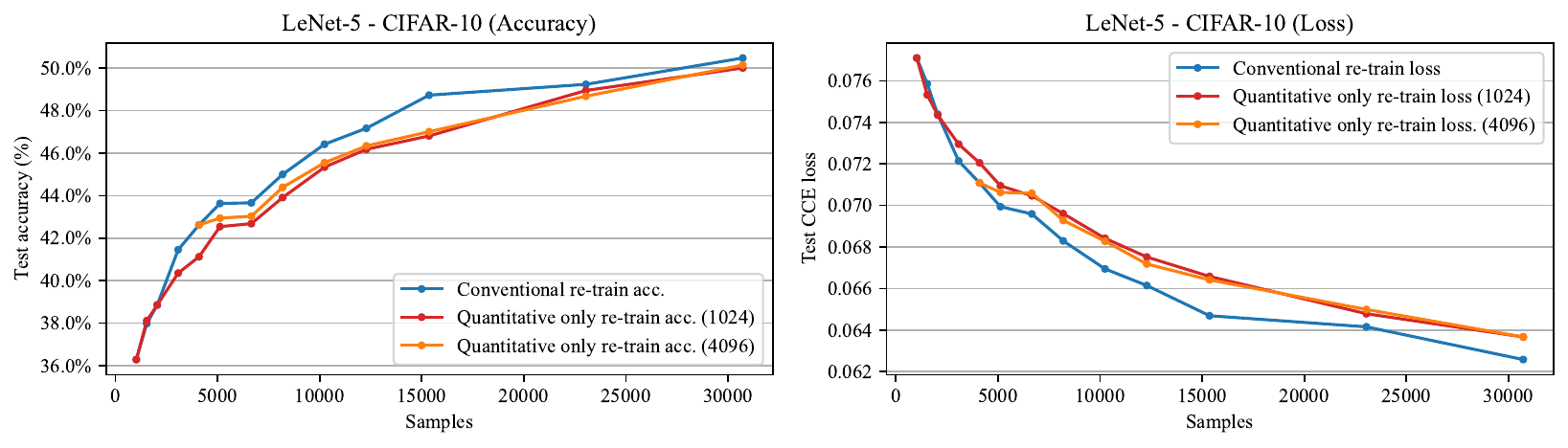}\\[0.1cm]
\includegraphics[width=1\columnwidth]{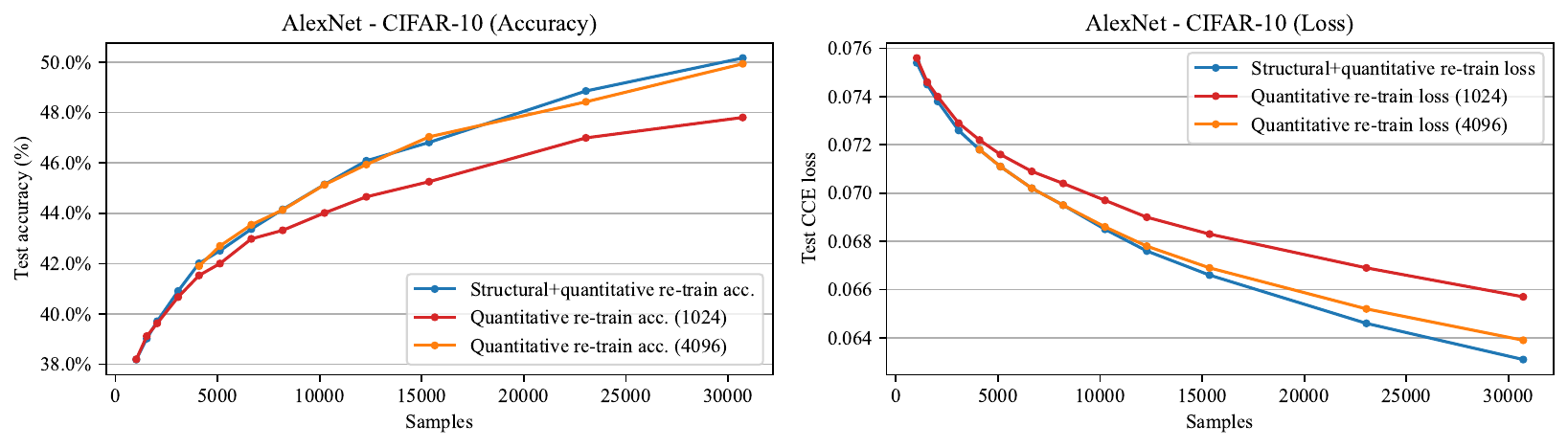}\\[0.1cm]
\includegraphics[width=1\columnwidth]{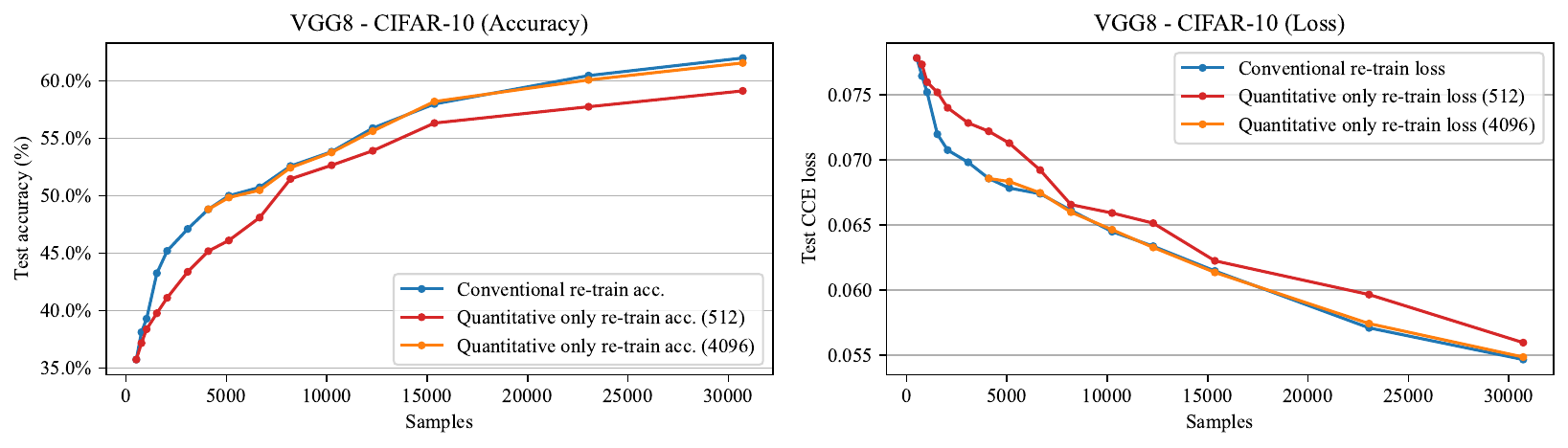}
\caption{Validation accuracy (left column) and CCE loss (right column) with traditional SGD and quantitative-only re-training using LeNet-5 (top row), AlexNet (middle row), and VGG8 (bottom row) for the CIFAR-10 dataset.}
\label{fig:retraining-results}\vspace{-0.4cm}
\end{figure}

Figure~\ref{fig:retraining-results} illustrates the validation accuracy and categorical cross entropy (CCE) validation loss achieved for the re-training of all classification layers of LeNet-5 (top row), AlexNet (middle row), and VGG8 (bottom row) using two algorithms: 1) Traditional SGD (blue line); and 2) SGD-based quantitative knowledge-only (red and orange lines).

For LeNet-5, as new samples are added in increments of 1,024 (from 2,048 to 32,768), the accuracy improvement rate is slightly slower when only the quantitative information is re-trained. Notably, after 8,192 samples, the accuracy for quantitative-only re-training declines at a higher rate. This behaviour was expected since the quantitative-only approach relied on fixed structural knowledge obtained from a small number of samples (1,024). However, a significant insight emerged from this experiment: the quantitative-only re-training knowledge continues to increase with the addition of more samples. Moreover, the CCE validation loss for both traditional SGD and quantitative knowledge-only re-training, using structural knowledge obtained from 1,024 samples, exhibits close similarity. This confirms the similar behaviour of the two systems. It is worth noting that the experiment was repeated using different random seeds for batch sample generation, yielding consistent results.

Regarding AlexNet, as new samples are added (from 1,024 to 32,768), the accuracy improves at a slightly slower rate when only the quantitative information is re-trained. This was expected since the quantitative-only approach relied on fixed structural knowledge obtained from a small number of samples (1,024). However, when the structural knowledge is re-initialised using 4,096 samples, the accuracy achieved is comparable to that of the traditional SGD algorithm. Nevertheless, it is important to note that the experiment that only uses 1,024 samples to obtain the structural knowledge still improves with the re-training step, even if the accuracy line falls below of the traditional SGD algorithm. The plots of CCE validation loss for both traditional SGD and quantitative knowledge-only re-training, using structural knowledge obtained from 1,024 samples, exhibits similar behaviour.

Similar behaviour is observed for the VGG8 model. However, due to its more complex architecture, the structural knowledge stabilises with 512 training samples. After this point, the quantitative-only re-training exhibits slightly slower convergence compared to the traditional SGD-based approach. While positive results are obtained, re-initialisation of structural knowledge may be necessary to address obsolete knowledge. Notably, re-training with 4,096 samples produces nearly similar results compared to the traditional SGD-based approach. The effects observed for both accuracy convergence and CCE validation loss are very similar.

In summary, the conducted experiments have demonstrated the feasibility and effectiveness of the proposed new training system. The utilisation of the path selector module and the estimator together has improved the performance of re-trained neural networks compared to traditional training methods. This advancement opens up new possibilities for optimising the training process and achieving better results in the application of neural networks across various domains and problems.

%% file: figures/lenet5.tex
  \tikzset{conv/.style={black,draw=conv,thick,fill=conv!20!white,rectangle,rounded corners=2.5,minimum height=0.8cm}}
  \begin{tikzpicture}
  \matrix [column sep=0.3cm]
  {
    \node[input,rotate=90,minimum width=3.7cm] (input) {\small\textsf{Input}}; &
    \node[conv,rotate=90,minimum width=3.7cm,align=center] (conv1) {\small \conv \tiny$(5\times 5\times 16)$ \\ \relu}; &
    \node[pool,rotate=90,minimum width=3.7cm] (pool1) {\small\apool \tiny$(2\times 2)$}; &
    \node[conv,rotate=90,minimum width=3.7cm,align=center] (conv2) {\small \conv \tiny$(5\times 5\times 6)$ \\ \relu}; &
    \node[pool,rotate=90,minimum width=3.7cm] (pool2) {\small\apool \tiny$(2\times 2)$}; &
    \node[fc,rotate=90,minimum width=3.7cm,align=center] (fc1) {\small\fc \tiny$(120)$ \\ \relu}; &
    \node[fc,rotate=90,minimum width=3.7cm,align=center] (fc2) {\small\fc \tiny$(84)$ \\ \relu}; &
    \node[fc,rotate=90,minimum width=3.7cm] (fc3) {\small\fc \tiny$(10)$}; &
    \node[sm,rotate=90,minimum width=3.7cm] (softmax) {\small\sm \tiny$(10)$}; &
    \node[input,rotate=90,minimum width=3.7cm] (output) {\small\textsf{Output}}; \\   
  };
    \draw[->] (input) -- (conv1) node[anchor=west,rotate=90,pos=0.5] {\scriptsize$32 \times 32 \times 3$};
    \draw[->] (conv1) -- (pool1) node[anchor=west,rotate=90,pos=0.5] {\scriptsize$32 \times 32 \times 6$};
    \draw[->] (pool1) -- (conv2) node[anchor=west,rotate=90,pos=0.5] {\scriptsize$16 \times 16 \times 6$};
    \draw[->] (conv2) -- (pool2) node[anchor=west,rotate=90,pos=0.5] {\scriptsize$12 \times 12 \times 16$};
    \draw[->] (pool2) -- (fc1)   node[anchor=west,rotate=90,pos=0.5] {\scriptsize$6 \times 6 \times 16$};
    \draw[->] (fc1) -- (fc2)     node[anchor=west,rotate=90,pos=0.5] {\scriptsize$120$};    
    \draw[->] (fc2) -- (fc3)     node[anchor=west,rotate=90,pos=0.5] {\scriptsize$84$};    
    \draw[->] (fc3) -- (softmax) node[anchor=west,rotate=90,pos=0.5] {\scriptsize$10$};    
    \draw[->] (softmax) -- (output) node[anchor=west,rotate=90,pos=0.5] {\scriptsize$10$};     
  \end{tikzpicture}

%% file: figures/alexnet.tex
  \tikzset{conv/.style={black,draw=conv,thick,fill=conv!20!white,rectangle,rounded corners=2.5,minimum height=0.8cm}}
  \begin{tikzpicture}
  \matrix [column sep=0.3cm]
  {
    \node[input,rotate=90,minimum width=3.7cm] (input) {\small\textsf{Input}}; &
    \node[conv,rotate=90,minimum width=3.7cm,align=center] (conv1) {\small \conv \tiny$(11\times 11\times 96)$ \\ \relu \\ \bn}; &
    \node[pool,rotate=90,minimum width=3.7cm] (pool1) {\small\mpool \tiny$(3\times 3)$}; &
    \node[conv,rotate=90,minimum width=3.7cm,align=center] (conv2) {\small \conv \tiny$(5\times 5\times 256)$ \\ \relu \\ \bn}; &
    \node[pool,rotate=90,minimum width=3.7cm] (pool2) {\small\mpool \tiny$(2\times 2)$}; &
    \node[conv,rotate=90,minimum width=3.7cm,align=center] (conv3) {\small \conv \tiny$(3\times 3\times 384)$ \\ \relu \\ \bn}; &
    \node[conv,rotate=90,minimum width=3.7cm,align=center] (conv4) {\small \conv \tiny$(3\times 3\times 384)$ \\ \relu \\ \bn}; &
    \node[conv,rotate=90,minimum width=3.7cm,align=center] (conv5) {\small \conv \tiny$(3\times 3\times 256)$ \\ \relu \\ \bn}; &
    \node[pool,rotate=90,minimum width=3.7cm] (pool3) {\small\mpool \tiny$(3\times 3)$}; &
    \node[fc,rotate=90,minimum width=3.7cm,align=center] (fc1) {\small\fc \tiny$(4096)$ \\ \relu}; &
    \node[fc,rotate=90,minimum width=3.7cm,align=center] (fc2) {\small\fc \tiny$(4096)$ \\ \relu}; &
    \node[fc,rotate=90,minimum width=3.7cm] (fc3) {\small\fc \tiny$(10)$}; &
    \node[sm,rotate=90,minimum width=3.7cm] (softmax) {\small\sm \tiny$(10)$}; &
    \node[input,rotate=90,minimum width=3.7cm] (output) {\small\textsf{Output}}; \\   
  };

    \draw[->] (input) -- (conv1) node[anchor=west,rotate=90,pos=0.5] {\scriptsize$32 \times 32 \times 3$};
    \draw[->] (conv1) -- (pool1) node[anchor=west,rotate=90,pos=0.5] {\scriptsize$32 \times 32 \times 96$};
    \draw[->] (pool1) -- (conv2) node[anchor=west,rotate=90,pos=0.5] {\scriptsize$16 \times 16 \times 96$};
    \draw[->] (conv2) -- (pool2) node[anchor=west,rotate=90,pos=0.5] {\scriptsize$12 \times 12 \times 256$};
    \draw[->] (pool2) -- (conv3) node[anchor=west,rotate=90,pos=0.5] {\scriptsize$12 \times 12 \times 256$};
    \draw[->] (conv3) -- (conv4) node[anchor=west,rotate=90,pos=0.5] {\scriptsize$6 \times 6 \times 384$};
    \draw[->] (conv4) -- (conv5) node[anchor=west,rotate=90,pos=0.5] {\scriptsize$6 \times 6 \times 384$};
    \draw[->] (conv5) -- (pool3)   node[anchor=west,rotate=90,pos=0.5] {\scriptsize$6 \times 6 \times 256$};    
    \draw[->] (pool3) -- (fc1)   node[anchor=west,rotate=90,pos=0.5] {\scriptsize$6 \times 6 \times 256$};    
    \draw[->] (fc1) -- (fc2)     node[anchor=west,rotate=90,pos=0.5] {\scriptsize$4096$};    
    \draw[->] (fc2) -- (fc3)     node[anchor=west,rotate=90,pos=0.5] {\scriptsize$4096$};    
    \draw[->] (fc3) -- (softmax) node[anchor=west,rotate=90,pos=0.5] {\scriptsize$10$};    
    \draw[->] (softmax) -- (output) node[anchor=west,rotate=90,pos=0.5] {\scriptsize$10$};   
  \end{tikzpicture}

%% file: figures/vgg8.tex
  \tikzset{conv/.style={black,draw=conv,thick,fill=conv!20!white,rectangle,rounded corners=2.5,minimum height=0.8cm}}
  \begin{tikzpicture}
  \matrix [column sep=0.3cm]
  {
    \node[input,rotate=90,minimum width=3.7cm] (input) {\small\textsf{Input }}; &
    \node[conv,rotate=90,minimum width=3.7cm,align=center] (conv1) {\small \conv \tiny$(3\times 3\times 64)$ \\ \relu \\ \bn}; &
    \node[pool,rotate=90,minimum width=3.7cm] (pool1) {\small\mpool \tiny$(2\times 2)$}; &
    \node[conv,rotate=90,minimum width=3.7cm,align=center] (conv2) {\small \conv \tiny$(3\times 3\times 128)$ \\ \relu \\ \bn}; &
    \node[pool,rotate=90,minimum width=3.7cm] (pool2) {\small\mpool \tiny$(2\times 2)$}; &
    \node[conv,rotate=90,minimum width=3.7cm,align=center] (conv3) {\small \conv \tiny$(3\times 3\times 256)$ \\ \relu \\ \bn}; &
    \node[pool,rotate=90,minimum width=3.7cm] (pool3) {\small\mpool \tiny$(2\times 2)$}; &
    \node[conv,rotate=90,minimum width=3.7cm,align=center] (conv4) {\small \conv \tiny$(3\times 3\times 512)$ \\ \relu \\ \bn}; &
    \node[pool,rotate=90,minimum width=3.7cm] (pool4) {\small\mpool \tiny$(2\times 2)$}; &
    \node[conv,rotate=90,minimum width=3.7cm,align=center] (conv5) {\small \conv \tiny$(3\times 3\times 512)$ \\ \relu \\ \bn}; &
    \node[pool,rotate=90,minimum width=3.7cm] (pool5) {\small\mpool \tiny$(2\times 2)$}; &
    \node[fc,rotate=90,minimum width=3.7cm,align=center] (fc1) {\small\fc \tiny$(4096)$ \\ \relu}; &
    \node[fc,rotate=90,minimum width=3.7cm,align=center] (fc2) {\small\fc \tiny$(4096)$ \\ \relu}; &
    \node[fc,rotate=90,minimum width=3.7cm] (fc3) {\small\fc \tiny$(10)$}; &
    \node[sm,rotate=90,minimum width=3.7cm] (softmax) {\small\sm \tiny$(10)$}; &
    \node[input,rotate=90,minimum width=3.7cm] (output) {\small\textsf{Output}}; \\   
  };

    \draw[->] (input) -- (conv1) node[anchor=west,rotate=90,pos=0.5] {\scriptsize$32 \times 32 \times 3$};
    \draw[->] (conv1) -- (pool1) node[anchor=west,rotate=90,pos=0.5] {\scriptsize$32 \times 32 \times 64$};
    \draw[->] (pool1) -- (conv2) node[anchor=west,rotate=90,pos=0.5] {\scriptsize$16 \times 16 \times 64$};
    \draw[->] (conv2) -- (pool2) node[anchor=west,rotate=90,pos=0.5] {\scriptsize$16 \times 16 \times 128$};
    \draw[->] (pool2) -- (conv3) node[anchor=west,rotate=90,pos=0.5] {\scriptsize$8 \times 8 \times 128$}; 
    \draw[->] (conv3) -- (pool3) node[anchor=west,rotate=90,pos=0.5] {\scriptsize$8 \times 8 \times 128$};
    \draw[->] (pool3) -- (conv4) node[anchor=west,rotate=90,pos=0.5] {\scriptsize$4 \times 4 \times 128$};    
    \draw[->] (conv4) -- (pool4) node[anchor=west,rotate=90,pos=0.5] {\scriptsize$4 \times 4 \times 128$};
    \draw[->] (pool4) -- (conv5) node[anchor=west,rotate=90,pos=0.5] {\scriptsize$2 \times 2 \times 128$};
    \draw[->] (conv5) -- (pool5) node[anchor=west,rotate=90,pos=0.5] {\scriptsize$2 \times 2 \times 128$};
    \draw[->] (pool5) -- (fc1)   node[anchor=west,rotate=90,pos=0.5] {\scriptsize$1 \times 1 \times 128$};
    \draw[->] (fc1) -- (fc2)     node[anchor=west,rotate=90,pos=0.5] {\scriptsize$4096$};    
    \draw[->] (fc2) -- (fc3)     node[anchor=west,rotate=90,pos=0.5] {\scriptsize$4096$};    
    \draw[->] (fc3) -- (softmax) node[anchor=west,rotate=90,pos=0.5] {\scriptsize$10$};    
    \draw[->] (softmax) -- (output) node[anchor=west,rotate=90,pos=0.5] {\scriptsize$10$};   
  \end{tikzpicture}

%% file: s6-greenlightningai.tex
\section{\glai System}\label{sec:greenlightningai}
In this section, we introduce and describe \glai, our new AI system that preserves structural knowledge while updating only quantitative knowledge with each re-training step.

\subsection{Definition}
\glai is a new AI system design that consists of a linear model capable of emulating the piece-wise linear behaviour of ReLU-based deep neural networks by subsetting the model for each particular sample. This AI system separately stores the information required to select the model subset for a given sample from the linear model parameters. By doing so, it makes it very easy to preserve the structural knowledge while re-training only the quantitative information. 

\begin{figure}[ht!]
\centering
\resizebox{0.8\textwidth}{!}{\input{figures/eeais}}\vspace{-0.1cm}
\caption{System diagram of the conceptual structure of the proposed AI system, including the path selector and the estimator.}
\label{fig:aisystem}
\end{figure}
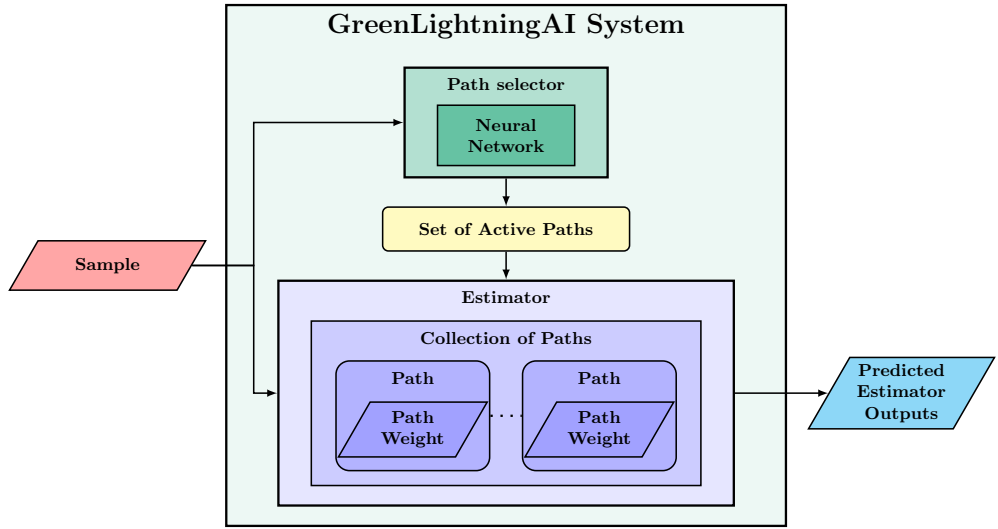

Figure~\ref{fig:aisystem} shows a diagram of \glai with the specific components (or parts) for the two decoupled knowledge classes. We reuse the terminology already introduced in the proof-of-concept AI system. The dataflow for inference is as follows: When a sample is introduced into the system, it passes through the path selector to obtain the set of active paths for that sample. This information is then passed to the estimator, which only activates the path weights associated with the active paths. To obtain the predicted output, the estimator multiplies the input sample components by the corresponding path weights and adds these products.

As opposed to the proof-of-concept AI system, the estimator module in \glai is defined as a single layer model with only linear relationships. This modified design does not change the functionality with respect to the proof-of-concept AI system design because multiple consecutive linear layers can be easily converted into a single linear layer just by expanding the mathematical expressions. This single-layer design facilitates the management of multiple copies of the model, each one trained with different datasets, and allows for easy combination of two or more copies just by computing a parameter-wise weighted average. As a result, this setup has the potential to reduce the cost, execution time, and energy consumption associated with re-training by several orders of magnitude. In particular, it enables incremental re-training by processing only the new samples. Despite this, it does not suffer from catastrophic forgetting because the old and new copies of the model can be easily merged as mentioned above. Model merging also enables other applications like federated learning and incremental federated learning.

The adoption of a single layer not only contributes a significant cost reduction, but it also increases flexibility by assigning independent weights to each path. In conventional neural network architectures, the path weights are determined by multiplying the weights along the corresponding path. Since paths partially overlap, they share connections and thus, path weights cannot be made independent of each other. However, the use of a single layer estimator in the \glai system promotes enhanced learning capabilities by eliminating those dependencies among paths.

\subsection{Implementation}
In this subsection, we describe in some detail the operation of \glai when first initialised and next re-trained multiple times with collections of new samples. Once a suitable neural network architecture has been chosen for the path selector, the workflow designed for this AI system can be split into the following four phases:

\begin{itemize}
\itemsep 0.1cm    
\item \textbf{Phase 1: Initial training of the path selector neural network.}
In this phase, the neural network chosen for the path selector is trained using a traditional method like SGD in order to acquire enough structural knowledge for the following phases. In general, this initial training requires much less samples and epochs than training a conventional neural network because the required accuracy is much lower. In particular, this neural network does not need to deliver accurate output results. Just the path activation status for each sample is required, and therefore, computation results are rounded to binary values. This phase corresponds to Phase 1 of the proof-of-concept (see  Section~\ref{phase1}).

\item \textbf{Phase 2: Initialising the path selector and estimator.}
Similarly to the proof-of-concept, the path selector of the \glai essentially leverages the trained neural network, directly using it to compute the set of active paths for each sample. The trained neural network is analysed to determine the number of paths it contains, and the estimator is defined as a single layer model that contains as many path weights as paths exist in the trained neural network. These path weights are initialised by multiplying the neuron weights along the corresponding path. During the re-training process, the path weights are the trainable parameters.

\item \textbf{Phase 3: Preparation of re-training: Obtaining the sets of active paths from the path selector.}
In this phase, each sample is passed through the path selector to obtain the set of active paths (i.e., the activation pattern) that will later be used by the estimator. This phase proceeds in the same way as phase 3 of the proof of concept (see section~\ref{phase1}).

\item \textbf{Phase 4: Re-training the estimator.}
The estimator processes each sample, taking into account the set of active paths delivered by the path selector in the previous phase. The forward pass of the re-training process calculates the predicted outputs as follows:
\begin{equation}\label{eq:output_estimator}
    o_{j} = \sum_{i=1}^{I} \sum_{k=1}^{N} \begin{cases}
        pw_{i,j,k} \cdot i_i & \text{if\ \ $p_{i,k}$ is active},\\
        0 & \text{otherwise}
        \end{cases}
        + 
        \sum_{k=1}^{M} \begin{cases}
        pw_{j,k} & \text{if\ \ $bp_{k}$ is active},\\
        0 & \text{otherwise}
    \end{cases}
\end{equation}
where $I$, $N$, and $M$ respectively represent the number of inputs, the number of paths ($p$) connecting each input ($i$) to each output ($o$), and the number of bias paths ($bp$). Also, $pw$ represents the path weight, regardless of whether it is a full path or a bias path. The predicted output is obtained by adding the contributions of all the active paths.

The backward pass uses the appropriate loss function to propagate its derivative back to the previous layer. Then, using the conventional chain rule, it updates the estimator path weights via SGD.
\end{itemize}

Compared to conventional retraining methods for DNNs, the fact that \glai decouples structural and quantitative knowledge and keeps the structural knowledge constant has a number of advantages:
\begin{itemize}
    \item The estimator consists of a single layer neural network, so it can be trained much faster than a DNN, avoiding the problems associated with the vanishing gradient (thanks to implementing a single layer), and even allowing the use of higher learning rates.
    \item As a linear system, it can be re-trained by processing only the new samples, later combining the resulting model with the model from the previous re-training step. This combination is equivalent to using the entire dataset, which avoids the problems associated with catastrophic forgetting. Note that model merging is possible because the path selector is kept constant. Thus, only the linear component of the AI system is retrained.
    \item As a linear system, the loss function error can be minimised by using direct solvers via QR or LU decomposition. Using direct solvers avoids potential stagnation in local minima inherent in iterative solvers.
\end{itemize}

In general, these four phases describe the re-training process of \glai, allowing for a more efficient re-training process compared to traditional SGD-based methods, where both structural and quantitative knowledge are learned simultaneously. Note that although phases 3 and 4 have been separately described for each sample, samples are processed in batches during re-training, as usual.

%% file: figures/eeais.tex
\tikzset{trape/.style={trapezium, draw,trapezium left angle=-120, trapezium right angle=-60}}

\begin{tikzpicture}

\node [draw,black,minimum width=10.2cm,minimum height=9.5cm,fill=green!60!blue!7,very thick,label={[yshift=-0.75cm]\Large\textbf{GreenLightningAI System}}] (base) {};

\node [draw,black,minimum width=3.7cm,minimum height=2.0cm,fill=green!60!blue!30,below=of base,yshift=9.4cm,label={[yshift=-0.55cm]\textbf{Path selector}},very thick] (pselector) {};

\node [draw,black,minimum width=2.5cm,minimum height=1.1cm,fill=green!60!blue!60,align=center,below=of pselector,yshift=2.35cm,thick,align=center] (nn) {\textbf{Neural}\\ \textbf{Network}};

\node [draw,black,minimum width=4.5cm,minimum height=0.8cm,fill=yellow!30!white,rounded corners=3,below=of pselector,yshift=14,thick] (sactive) {\textbf{Set of Active Paths}};

\node [draw,black,minimum width=8.3cm,minimum height=4.1cm,fill=blue!10!white,below=of sactive,yshift=14,label={[yshift=-0.55cm]\textbf{Estimator}},very thick] (estimator) {};

\node [draw,black,minimum width=7.1cm,minimum height=3.0cm,fill=blue!20!white,yshift=4.4cm,thick,align=center,below=of estimator,label={[yshift=-0.55cm]{\textbf{Collection of Paths}}}] (collection) {};

\node [draw,black,minimum width=2.8cm,minimum height=2.0cm,fill=blue!30!white,rounded corners=7,below=of collection,yshift=3.3cm,xshift=-1.7cm,thick,label={[yshift=-0.55cm]{\textbf{Path}}}] (path1) {};

\node [trape,black,minimum width=2.3cm,minimum height=1cm,fill=blue!37!white,below=of collection,yshift=2.55cm,xshift=-1.7cm,thick,align=center] (pathw1) {\textbf{Path}\\ \textbf{Weight}};

\node [draw,black,minimum width=2.8cm,minimum height=2.0cm,fill=blue!30!white,rounded corners=7,below=of collection,yshift=3.3cm,xshift=+1.7cm,thick,label={[yshift=-0.55cm]{\textbf{Path}}}] (path2) {};

\node [trape,black,minimum width=2.3cm,minimum height=1cm,fill=blue!37!white,below=of collection,yshift=2.55cm,xshift=+1.7cm,thick,align=center] (pathw2) {\textbf{Path}\\ \textbf{Weight}};

\node [trape,black,minimum width=0.8cm,minimum height=0.9cm,fill=red!35!white,left=of base,thick,xshift=0.4cm] (sample) {\textbf{Sample}};
    
\node [trape,black,minimum width=3.4cm,minimum height=0.9cm,fill=cyan!40!white,right=of estimator,thick,xshift=0.7cm,align=center] (prediction) {\textbf{Predicted}\\ \textbf{Estimator}\\ \textbf{Outputs}};

\draw[->,thick] (pselector) -- (sactive);
\draw[->,thick] (sactive) -- (estimator);
\draw[->,thick] (sample.east) -- (-4.6,0) |- (pselector.west);
\draw[->,thick] (sample.east) -- (-4.6,0) |- (estimator.west);
\draw[->,thick] (estimator) -- (prediction);

\draw[-,loosely dotted,very thick] (path1) -- (path2);

\end{tikzpicture}

%% file: s7-remarks.tex
\section{Concluding Remarks}\label{sec:remarks}
In this work, we have presented \glai, a novel AI system that decouples structural from quantitative knowledge learning to improve its performance and reduce the energy consumption with respect to DNN re-training algorithms. The design of this system includes the path selector, which is able to determine the activation patterns for each sample, and the estimator, which is able to fine-tune the weights and biases of the active paths determined by the path selector for the samples in the dataset. 

Throughout the results, we have demonstrated the feasibility and effectiveness of this approach. The experimental evaluation using the LeNet-5, AlexNet and VGG8 models on the CIFAR-10 dataset has given us valuable insights into the behaviour and convergence of the re-trained networks. We observed several interesting results. First, we can confirm that during the initial training phase of the neural network, the structural knowledge converges into a stable state faster than the quantitative knowledge, and this structural knowledge is sufficient to generate activation patterns that keep the whole model convergence rate close to that obtained by the original SGD-based algorithm. Moreover, when re-training with only quantitative information, the accuracy improvement rate was slightly lower compared to the traditional SGD-based re-training. This can be attributed both to keeping structural knowledge constant and to having obtained it from a smaller number of samples. However, it is important to note that the quantitative-only approach showed continuous knowledge improvement as more samples were added, and that the CCE validation loss remained very close to that of traditional SGD. These results therefore demonstrate the high potential of using quantitative knowledge for efficient re-training.

Furthermore, the experiments revealed that re-initialising the structural knowledge at specific points, using a larger number of samples, leads to a model accuracy comparable to that of the traditional SGD-based approach. This finding emphasises the significance of updating the structural knowledge in line with the evolving data to achieve high accuracy.

As future work, we plan to explore several directions. First, we aim to investigate the application of our AI system and associated training methods to other network architectures and datasets to assess its generalisation capabilities. We also plan to analyse the scalability of this approach, measuring the re-training speedup with respect to conventional re-training for large datasets. In addition, we intend to explore adaptive strategies for updating the structural knowledge during the re-training phase to address the problem of outdated knowledge. We are also considering the design of shallow path selectors, and even path selectors that are not based on neural networks. By addressing these aspects, we expect to advance the field of neural network re-training and improve the performance and applicability of neural networks in different domains and problem settings.

%% file: greenlightningai.bbl

%% file: greenlightningai.bbl
\begin{thebibliography}{10}
\providecommand{\url}[1]{#1}
\csname url@samestyle\endcsname
\providecommand{\newblock}{\relax}
\providecommand{\bibinfo}[2]{#2}
\providecommand{\BIBentrySTDinterwordspacing}{\spaceskip=0pt\relax}
\providecommand{\BIBentryALTinterwordstretchfactor}{4}
\providecommand{\BIBentryALTinterwordspacing}{\spaceskip=\fontdimen2\font plus
\BIBentryALTinterwordstretchfactor\fontdimen3\font minus \fontdimen4\font\relax}
\providecommand{\BIBforeignlanguage}[2]{{%
\expandafter\ifx\csname l@#1\endcsname\relax
\typeout{** WARNING: IEEEtran.bst: No hyphenation pattern has been}%
\typeout{** loaded for the language `#1'. Using the pattern for}%
\typeout{** the default language instead.}%
\else
\language=\csname l@#1\endcsname
\fi
#2}}
\providecommand{\BIBdecl}{\relax}
\BIBdecl

\bibitem{amodei2018ai}
D.~Amodei and D.~Hernandez, ``Ai and compute,'' \url{https://openai.com/blog/ai-and-compute}, 2018, last accessed: April 2023.

\bibitem{nvidia2022h100}
N.~Corp., ``Nvidia h100 tensor core gpu architecture,'' \url{https://nvdam.widen.net/s/9bz6dw7dqr/gtc22-whitepaper-hopper}, 2022, last accessed: April 2023.

\bibitem{hennessy2019new}
J.~L. Hennessy and D.~A. Patterson, ``A new golden age for computer architecture,'' \emph{Communications of the ACM}, vol.~62, no.~2, pp. 48--60, 2019.

\bibitem{reuther2021ai}
A.~Reuther, P.~Michaleas, M.~Jones, V.~Gadepally, S.~Samsi, and J.~Kepner, ``Ai accelerator survey and trends,'' in \emph{2021 IEEE High Performance Extreme Computing Conference (HPEC)}.\hskip 1em plus 0.5em minus 0.4em\relax IEEE, 2021, pp. 1--9.

\bibitem{huang2006universal}
G.-B. Huang, L.~Chen, C.~K. Siew \emph{et~al.}, ``Universal approximation using incremental constructive feedforward networks with random hidden nodes,'' \emph{IEEE Trans. Neural Networks}, vol.~17, no.~4, pp. 879--892, 2006.

\bibitem{zhang2016comprehensive}
L.~Zhang and P.~N. Suganthan, ``A comprehensive evaluation of random vector functional link networks,'' \emph{Information sciences}, vol. 367, pp. 1094--1105, 2016.

\bibitem{cao2018review}
W.~Cao, X.~Wang, Z.~Ming, and J.~Gao, ``A review on neural networks with random weights,'' \emph{Neurocomputing}, vol. 275, pp. 278--287, 2018.

\bibitem{bianchi2017recurrent}
F.~M. Bianchi, E.~Maiorino, M.~C. Kampffmeyer, A.~Rizzi, and R.~Jenssen, ``Recurrent neural networks for short-term load forecasting: an overview and comparative analysis,'' 2017.

\bibitem{klabjan2020neural}
D.~Klabjan and X.~Zhu, ``Neural network retraining for model serving,'' 2020.

\bibitem{chatzis2018forecasting}
S.~P. Chatzis, V.~Siakoulis, A.~Petropoulos, E.~Stavroulakis, and N.~Vlachogiannakis, ``Forecasting stock market crisis events using deep and statistical machine learning techniques,'' \emph{Expert systems with applications}, vol. 112, pp. 353--371, 2018.

\bibitem{gerlein2016evaluating}
E.~A. Gerlein, M.~McGinnity, A.~Belatreche, and S.~Coleman, ``Evaluating machine learning classification for financial trading: An empirical approach,'' \emph{Expert Systems with Applications}, vol.~54, pp. 193--207, 2016.

\bibitem{ma2019financial}
X.~Ma and S.~Lv, ``Financial credit risk prediction in internet finance driven by machine learning,'' \emph{Neural Computing and Applications}, vol.~31, pp. 8359--8367, 2019.

\bibitem{almahdi2017adaptive}
S.~Almahdi and S.~Y. Yang, ``An adaptive portfolio trading system: A risk-return portfolio optimization using recurrent reinforcement learning with expected maximum drawdown,'' \emph{Expert Systems with Applications}, vol.~87, pp. 267--279, 2017.

\bibitem{sonksen2022machine}
J.~S{\"o}nksen, ``Machine learning for asset pricing,'' \emph{Econometrics with Machine Learning}, pp. 337--366, 2022.

\bibitem{kelley2018artificial}
K.~H. Kelley, L.~M. Fontanetta, M.~Heintzman, and N.~Pereira, ``Artificial intelligence: Implications for social inflation and insurance,'' \emph{Risk Management and Insurance Review}, vol.~21, no.~3, pp. 373--387, 2018.

\bibitem{dick2019deep}
K.~Dick, L.~Russell, Y.~Souley~Dosso, F.~Kwamena, and J.~R. Green, ``Deep learning for critical infrastructure resilience,'' \emph{Journal of Infrastructure Systems}, vol.~25, no.~2, p. 05019003, 2019.

\bibitem{tinn2023fine}
R.~Tinn, H.~Cheng, Y.~Gu, N.~Usuyama, X.~Liu, T.~Naumann, J.~Gao, and H.~Poon, ``Fine-tuning large neural language models for biomedical natural language processing,'' \emph{Patterns}, vol.~4, no.~4, 2023.

\bibitem{french1999catastrophic}
R.~M. French, ``Catastrophic forgetting in connectionist networks,'' \emph{Trends in Cognitive Sciences}, vol.~3, no.~4, pp. 128--135, April 1999.

\bibitem{kirkpatrick2017overcoming}
J.~Kirkpatrick, R.~Pascanu, N.~Rabinowitz, J.~Veness, G.~Desjardins, A.~A. Rusu, K.~Milan, J.~Quan, T.~Ramalho, A.~Grabska-Barwinska \emph{et~al.}, ``Overcoming catastrophic forgetting in neural networks,'' \emph{Proceedings of the national academy of sciences}, vol. 114, no.~13, pp. 3521--3526, 2017.

\bibitem{rolnick2019experience}
D.~Rolnick, A.~Ahuja, J.~Schwarz, T.~Lillicrap, and G.~Wayne, ``Experience replay for continual learning,'' \emph{Advances in Neural Information Processing Systems}, vol.~32, 2019.

\bibitem{van2020brain}
G.~M. Van~de Ven, H.~T. Siegelmann, and A.~S. Tolias, ``Brain-inspired replay for continual learning with artificial neural networks,'' \emph{Nature communications}, vol.~11, no.~1, p. 4069, 2020.

\bibitem{yu2018slimmable}
J.~Yu, L.~Yang, N.~Xu, J.~Yang, and T.~Huang, ``Slimmable neural networks,'' \emph{arXiv preprint arXiv:1812.08928}, 2018.

\bibitem{rusu2016progressive}
A.~A. Rusu, N.~C. Rabinowitz, G.~Desjardins, H.~Soyer, J.~Kirkpatrick, K.~Kavukcuoglu, R.~Pascanu, and R.~Hadsell, ``Progressive neural networks,'' \emph{arXiv preprint arXiv:1606.04671}, 2016.

\bibitem{auda1999modular}
G.~Auda and M.~Kamel, ``Modular neural networks: a survey,'' \emph{International journal of neural systems}, vol.~9, no.~02, pp. 129--151, 1999.

\bibitem{chen2015deep}
K.~Chen, ``Deep and modular neural networks,'' \emph{Springer Handbook of Computational Intelligence}, pp. 473--494, 2015.

\bibitem{castillo2002global}
E.~Castillo, O.~Fontenla-Romero, B.~Guijarro-Berdinas, and A.~Alonso-Betanzos, ``A global optimum approach for one-layer neural networks,'' \emph{Neural Computation}, vol.~14, no.~6, pp. 1429--1449, 2002.

\bibitem{Hartmann21}
\BIBentryALTinterwordspacing
D.~Hartmann, D.~Franzen, and S.~Brodehl, ``Studying the evolution of neural activation patterns during training of feed-forward relu networks,'' \emph{Frontiers in Artificial Intelligence}, vol.~4, 2021. [Online]. Available: \url{https://www.frontiersin.org/articles/10.3389/frai.2021.642374}
\BIBentrySTDinterwordspacing

\bibitem{PyDTNN2}
S.~Barrachina, A.~Castelló, M.~Catalán, M.~F. Dolz, and J.~I. Mestre, ``A flexible research-oriented framework for distributed training of deep neural networks,'' in \emph{2021 IEEE International Parallel and Distributed Processing Symposium Workshops (IPDPSW)}, 2021, pp. 730--739.

\end{thebibliography}
